\documentclass[10pt,journal,compsoc]{IEEEtran}

\ifCLASSOPTIONcompsoc
  \usepackage[nocompress]{cite}
\else
  \usepackage{cite}
\fi
\ifCLASSINFOpdf
  \usepackage[pdftex]{graphicx}
  \graphicspath{{../pdf/}{../jpeg/}}
  \DeclareGraphicsExtensions{.pdf,.jpeg,.png}
\else
\fi
\usepackage{color}

\usepackage{amsmath}
\usepackage{amsmath,amsfonts,amssymb}

\usepackage{mdwtab}
\newcommand*{\mutual}{\ensuremath{m}}
\newcommand*{\exclusive}{\ensuremath{e}}
\newcommand*{\defaultlayerdim}{\ensuremath{200}}
\newcommand*{\vision}{\ensuremath{v}}
\newcommand*{\proprioception}{\ensuremath{p}}
\newcommand*{\preencoding}{{\ensuremath{\text{pre}}}}
\newcommand*{\postencoding}{{\ensuremath{\text{post}}}}
\newcommand*{\position}{{\ensuremath{\text{pos}}}}
\newcommand*{\velocity}{{\ensuremath{\text{vel}}}}
\newcommand*{\eeff}{{\ensuremath{\text{ee}}}}
\newcommand*{\target}{{\ensuremath{\text{target}}}}
\newcommand*{\readout}{{\ensuremath{\text{readout}}}}

\newcommand*{\revisioncolor}{}

\newcommand*{\charlescolor}{}

\begin{document}
%
\title{Learning Abstract Representations through Lossy Compression of Multi-Modal Signals}
%
%
%
%

\author{Charles~Wilmot, Gianluca~Baldassarre, and Jochen~Triesch,~\IEEEmembership{Member,~IEEE}

\IEEEcompsocitemizethanks{\IEEEcompsocthanksitem C.~Wilmot and J.~Triesch are with the Frankfurt Institute for Advanced Studies, Ruth-Moufang-Str.~1, 60438 Frankfurt am Main,
Germany. \{wilmot,triesch\}\@fias.uni-frankfurt.de
\IEEEcompsocthanksitem G.~Baldassarre is with the National Research Council, Institute of Cognitive Sciences and Techologies, Via S. Martino della Battaglia 44, I-00185 Rome, Italy. gianluca.baldassarre@istc.cnr.it \protect\\
}
\thanks{Manuscript received September 30, 2020; revised September 31, 2020.}}

%
%

\markboth{IEEE TRANSACTIONS ON COGNITIVE AND DEVELOPMENTAL SYSTEMS}%
{Shell \MakeLowercase{\textit{et al.}}: Bare Advanced Demo of IEEEtran.cls for IEEE Computer Society Journals}
%



\maketitle

\begin{abstract}
A key competence for open-ended learning is the formation of increasingly abstract representations useful for driving complex behavior. Abstract representations ignore specific details and facilitate generalization. Here we consider the learning of abstract representations in a multi-modal setting with two or more input modalities. We treat the problem as a lossy compression problem and show that generic lossy compression of multimodal sensory input naturally extracts abstract representations that tend to strip away modalitiy specific details and preferentially retain information that is shared across the different modalities.
{\revisioncolor 
Specifically, we propose an architecture that is able to extract information common to different modalities based on the compression abilities of generic autoencoder neural networks. We test the architecture with two tasks that allow 1) the precise manipulation of the amount of information contained in and shared across different modalities and 2) testing the method on a simulated robot with visual and proprioceptive inputs. Our results show the validity of the proposed approach and demonstrate the applicability to embodied agents.}

\end{abstract}



%

\section{Introduction}
\label{sec:introduction}

%
%
%
%
\IEEEPARstart{H}{uman} intelligence rests on the ability to learn abstract representations. An abstract representation has stripped away many details of specific examples of a concept and retains what is common, {\revisioncolor thereby facilitating generalization and transfer of knowledge to new tasks \cite{WeissKhoshgoftaarWang2016}.} A key challenge for natural and artificial developing agents is to learn such abstract representations. How can this be done?

\subsection{Learning abstract representations}

In classic {\em supervised learning} settings, an external teacher provides the abstract concept by virtue of explicit labeling of training examples. For example, in neural network based image recognition explicit labels ({\em cat, dog,} etc.) are provided as a ``one-hot'' abstract code and the network learns to map input images to this abstract representation \cite{supervised1, supervised2, supervised3}. While such a learned mapping qualifies as an abstract representation that has 
``stripped away many details of specific examples of a concept and retains what is common'' it needs to be provided by the teacher through typically millions of examples. This is clearly not how human infants learn and often leads to undesirable generalization behavior \cite{one_pixel_attack, adversarial_patch}. Therefore unsupervised learning and reinforcement learning are more interesting settings for studying the autonomous formation of abstract representations.

{\em Reinforcement learning} (RL) can also be viewed from the perspective of learning abstract representations. The essence of RL is to learn a {\em policy}, i.e., a mapping from states of the world to actions that an agent should take in order to maximize the sum of future rewards \cite{sutton}. Often, this mapping is realized through neural networks. In deep Q-learning networks \cite{dqn, double_dqn}, for example, a neural network learns to map the current state of the world (e.g., the current image of a computer game screen) onto the expected future rewards for performing different actions (e.g., joystick commands) in this particular state. This can be interpreted as the agent learning an abstract ``concept'' of the following kind: {\em the set of all world states for which this particular joystick command will be the optimal choice of action}. While these ``concepts'' are not provided {\em explicitly} by a teacher, they are provided {\em implicitly} through the definition of the RL problem (states, actions, rewards, environment dynamics). In fact, from an RL perspective, these ``concepts'' are the {\em only} ones the agent ever needs to know about. They suffice for behaving optimally in this particular environment. However, they may also become completely useless when the task changes, e.g., a different computer game should be played. This exemplifies the deep and unresolved problem of how abstract knowledge could be extracted in RL that is likely to transfer to new tasks. 

In the domain of {\em unsupervised learning}, on which we will focus in the remainder, a simple approach to learning somewhat abstracted representations is through {\em clustering}. For example, in $k$-means clustering \cite{kmean} an input $x \in \mathbb{R}^n$ is represented by mapping it to one of $k$ cluster centers $ c_i \in \mathbb{R}^n, \, i \in \{1, \ldots, k\} $ based on a suitably defined distance metric in $\mathbb{R}^n$. Representing an input $x$ by the closest cluster center strips away information about the precise location of $x$ in $\mathbb{R}^n$, achieving a simple form of abstraction. However, the use of a predefined distance metric is limiting. A second set of approaches for learning more abstract representation through lossy compression are dimensionality reduction techniques. A classic example of such an approach is principal component analysis (PCA). PCA finds linear projections of the data such that the projected data has maximum variance, while orthogonal directions are discarded. Thus, the information that gets stripped away corresponds to linear projections of small variance. The central limitation of PCA is the restriction to linear projections. A popular and more powerful approach is the use of autoencoder neural networks \cite{autoencoder}, on which we will focus in the following. Like other dimensionality reduction techniques, autoencoders construct a more compact and abstract representation of the input domain by learning to map inputs $x \in \mathbb{R}^n$ onto a more compact latent representation $z \in \mathbb{R}^m$ with $m \ll n$ via a neural network. For this, the network has an encoder/decoder structure with a ``bottleneck'' in the middle. The $n$-dimensional input is mapped via several layers onto the $m$-dimensional bottleneck and from there via several layers to an $n$-dimensional output. The learning objective is to reconstruct the input at the output, but the trivial solution of a direct identity mapping is avoided, because the input needs to be ``squeezed through'' the central $m$-dimensional bottleneck. After training, the decoder is often discarded and only the encoder is retained, providing a mapping from the original $n$-dimensional input $x$ to a compressed lower-dimensional latent representation $z$. Due to the nonlinear nature of the neural network, it is difficult to characterize exactly what information will be stripped away in the encoding process. An exception is the special case of a linear network with a quadratic loss function. For this case, it can be shown that the network discovers the same subspace as linear PCA. In the following, we will consider the implications of a developing agent learning to encode an input $x$ that comprises multiple sensory modalities. 

\subsection{Multimodality, Abstraction, and Lossy Compression: An Information Theoretic Perspective}
Our different sensory modalities (vision, audition, proprioception, touch, smell, taste) provide us with different ``views'' of our physical environment. These views are not independent, but contain shared information about the underlying physical reality. Therefore, it must be possible to compress the information coming from the different modalities into a more compact code. As an example, consider viewing and touching your favorite coffee cup. Some information such as the color or the text or picture printed on the cup will only be accessible to the visual modality. Some information, such as the temperature or roughness of the surface, will only be accessible to the haptic modality. Some information, however, such as the 3-D shape of the cup, will be accessible to both modalities. This implies potential for compression. Let $X_v$ represent the visual input and $X_h$ the haptic input. We can quantify the amount of information using concepts from information theory. For now, we will ignore the fact that the amount of information is a function of our behavior, but we will return to this point in the Discussion. We can quantify the amount of information in $X_v$ and $X_h$ as:
\begin{equation}
    H(X_v, X_h) = H(X_v) + H(X_h) - MI(X_v; X_h) \; ,
\end{equation}
where $H(X_v$) and $H(X_h)$ are the individual entropies of the visual and haptic signals, respectively, $H(X_v, X_h)$ is their joint entropy, and $MI(X_v; X_h)$ is their mutual information, i.e., the amount of information that $X_v$ and $X_h$ have in common. The individual entropies $H(X_v$) and $H(X_h)$ and the joint entropy $H(X_v, X_h)$ indicate, respectively, how many bits are required on average to encode $X_v$ and $X_h$ individually or jointly. The mutual information expresses how many bits can be ``saved'' by jointly encoding $X_v$ and $X_h$ compared to encoding them separately. If the visual and haptic inputs were statistically independent, then $MI(X_v; X_h)=0$ and no savings can be gained. If there are any statistical dependencies between $X_v$ and $X_h$, then $MI(X_v; X_h)>0$ and $H(X_v, X_h) < H(X_v) + H(X_h)$, i.e., the visual and haptic signals can be compressed into a more compact code. In principle, this compression can be achieved without any loss of information.

However, it is the very nature of abstraction to ``strip away'' information about the details of a situation and only maintain a ``coarser'' and hopefully more generalizable representation. Consider again the situation of visuo-haptic perception of your favorite coffee cup. If we were to strip away the information that is only accessible to the visual modality (color and printed text/picture) and strip away the information that is accessible to only the haptic modality (temperature, surface roughness), then we are left with a much reduced and abstract representation that maintains essentially the 3-D shape of the cup and allows for many generalizations, e.g., how to grasp {\em this} particular cup versus many similarly shaped cups with virtually endless combinations of color, picture, text, surface roughness, and temperature. Thus, learning an abstract code by retaining information that is shared across modalities and stripping away information that is specific to only individual modalities may lead to very useful representations with high potential for generalization. How can this be done?

Here we investigate a possible solution that relies on autoencoding the multimodal inputs into a sufficiently compact lossy code. The rationale for this approach is as follows. Consider an autoencoder that learns to map the concatenation of $X_v$ and $X_h$ onto a small latent vector $Z$ such that $X_v$ and $X_h$ can be reconstructed from $Z$ with minimal loss. What information should $Z$ encode? If the information coding capacity of $Z$ is at least $H(X_v, X_h)$, then it can simply encode all the information contained in the visual and haptic signals. If it is smaller, however, then some information must be discarded. So what information should be kept and what information should be discarded? In general, it appears best to keep the information that $X_v$ and $X_h$ have in common, because this information aids in reconstructing both of them, essentially killing two (metaphorical) birds with one stone. Information that is present in only either $X_v$ or $X_h$ cannot be helpful in reconstructing the other. Therefore, an autoencoder with limited capacity will learn a representation that tends to prioritize the information that $X_v$ and $X_h$ have in common and tends to strip away the information that is unique to either modality. This is exactly the kind of abstract representation that we would like to achieve.

In the remainder of this article we make these ideas more concrete and study them in extensive computer simulations. {\revisioncolor We begin by a discussion of related work.} We then propose a number of models to learn abstract representations from multimodal input via autoencoding and compare their behavior. {\revisioncolor In a first approach, we use synthetically generated inputs, since this allows us to precisely control the amount of information in individual sensory modalities and the amount of information that they share. In a second approach, we use visual and proprioceptive data from a robot simulation.} We end by discussing broader implications of learning abstract representations through lossy compression of multimodal input for cognitive development.

\section{Related Work}
\label{sec:related-work}

{\revisioncolor
Our work falls in the general area of unsupervised representation learning \cite{BengioCourvilleVincent2013}. Representation learning aims to build machine learning algorithms that extract the explanatory variation factors of data. In so doing, the learned representations are also often compressed/abstracted in that they discard some information.
Algorithms for representation learning can be grouped based on the general strategy they use to form representations. This strategy plays the role of the choice of {\em generic priors} on the process that is assumed to have generated the data. These priors might for example assume that there are indeed distinct factors capturing the variability of data (e.g., changing positions of objects and light sources relative to a camera giving rise to varying camera images), that observations close in space or time have similar values (e.g., as in natural images and videos), or that most of the probability mass of the data concentrates on manifolds with a dimensionality smaller than that of the perceptual space (e.g., as assumed in autoencoders).

Information theoretic approaches to learning representations are of particular interest and have a long history. A complete review is beyond the scope of this article. Here we focus on relating our work to some classic approaches.

In {\em Efficient Coding} \cite{attneave1954some,barlow1961}, the goal is to learn a representation for sensory signals that is ``compact'' and exploits redundancies in the signals to arrive at a more efficient code. Often this is formulated as maximizing the mutual information $I(X;Y)$ between an input $X$ and its representation $Y$ while putting additional constraints on $Y$. For example, in linear and non-linear {\em independent component analysis} (ICA) \cite{hyvarinen2000independent}, one attempts to make the components of $Y$ statistically independent, corresponding to the prior of the data being generated by mixing independent information sources. Searching for codes that are factorial while retaining the maximum amount of information about the input leads to an efficient code. This is because if the extracted components $Y$ were not independent, then there would be potential for further compression of $Y$ to generate an even more efficient code.

{\em Sparse coding} is a popular variant of efficient coding that is related to ICA and imposes a sparsity constraint (or prior) on $Y$ \cite{olshausen1997sparse}. This has given rise to a large literature on learning sparse representations for sensory signals. Sparse coding models are frequently employed as models of learning sensory representations in the brain. More recently, such approaches have also been extended to active perception. In the Active Efficient Coding (AEC) framework, the sensory encoding is optimized by simultaneously optimizing the encoding of the sensory inputs and the movements of the sense organs that shape the statistics of these inputs \cite{zhao2012unified,eckmann2020active}. 

{\em Predictive coding} can be viewed as a special case of efficient coding that forms hierarchical representations, where higher levels predict the activity of lower levels and lower levels signal prediction errors to higher levels \cite{rao1999predictive}. 
Note that in these approaches the goal is generally to retain as much information about the input as possible There is no notion of abstraction as we have defined it, i.e., of deliberately discarding information to arrive at a representation that generalizes more easily to new situations. 

An alternative information theoretic learning objective was proposed by Becker and Hinton \cite{becker1992self}. In their IMAX approach, two distinct inputs are considered (in particular visual inputs from two neighboring locations of stereoscopic image pairs) and the objective is to extract information that is {\em shared} by these two input sources. Becker and Hinton demonstrate that this allows learning to extract disparity information from the stereoscopic images. While the specific texture projected to the two neighboring retinal locations may be different, the binocular disparity is typically the same, because it tends to vary smoothly across the image. Their approach can be viewed as an early attempt to extract more ``abstract'' information (disparity) by encoding information from multiple sources (binocular visual inputs from two neighboring locations) and only keeping the information that they have in common.

IMAX can also be viewed as related to Canonical Correlation Analysis (CCA) \cite{hotelling1936relations}. In CCA one tries to find linear combinations of the components of a random vector $X$ and the components of a random vector $Y$ such that these linear combinations are maximally correlated. A high correlation between the two projections implies that they share a substantial amount of information (correlation implies statistical dependence and therefore non-zero mutual information while the reverse is generally not true).

The information bottleneck method by Tishby and colleagues considers the objective of encoding a sensory input $X$ and only keeping the information that is useful for predicting a second ``relevant'' signal $Y$ \cite{tishby1999information}. Thus it also aims at keeping only information that is shared with (or: predictive of) another signal, while discarding all other information. However, in contrast to IMAX there is a clear asymmetry between the input $X$ and the signal $Y$. The information bottleneck objective can be expressed as follows: Let $T$ be a compressed version of $X$. $T$ should retain as little information from $X$ as possible, but at the same time keep as much information about $Y$ as possible. Thus $T$ functions as the information bottleneck. Formally, one is seeking to optimize the functional:
\begin{equation}
    \min_{p(T=t|X=x)} I(X;T) - \beta I(T;Y) ,
\end{equation}
where $p(T=t|X=x)$ describes the encoding of an input $x$ via its representation $t$, $I(X;T)$ is the mutual information between $X$ and $T$, $I(T;Y)$ is the mutual information between $T$ and $Y$, and $\beta$ is a parameter for balancing the two objectives. Unfortunately, as shown in the Appendix, it is not straightforward to derive a symmetric variant of the information bottleneck, i.e., to find an encoding $p(T=t|X=x,Y=y)$ that would keep only the information that $X$ and $Y$ share. 

Instead, in our approach we utilize the generic ability of (deep) autoencoders to perform lossy compression of sensory inputs. Autoencoders implicitly assume that much of the variability of the data can be accounted for by a smaller number of latent causes. Specifically, in our case we assume that inputs from different sensory modalities are related since they are consequences of the same underlying physical causes. For example, a human infant (or robot) hitting their hand on the table will observe the consequences in multiple modalities (feeling the contact, seeing the movement, hearing the sound). The choice of the size of the autoencoder's bottleneck is analogous to the choice of $\beta$ in the information bottleneck. Note that many different types of autoencoders exist including denoising autoencoders \cite{vincent2008extracting}, sparse autoencoders \cite{ng2011sparse}, contractive autoencoders \cite{rifai2011contractive}, variational autoencoders \cite{kingma2013auto}, adversarial autoencoders \cite{makhzani2015adversarial}, etc.  In order to not distract from our main point about lossy compression of multimodal signals, we restrict our experiments to ``generic'' autoencoders. It should be understood that the results are expected to generalize to other types of autoencoders or, indeed, other lossy compression schemes. The novelty of the proposed solution resides in the fact that it exploits the multimodal organisation of the information available to an agent to form abstract representations. Furthermore, we propose a specific cross-modality prediction architecture to distill only the information that is shared across multiple modalities.
}

%
%

\begin{figure}[!t]
\centering
\includegraphics[width=\columnwidth]{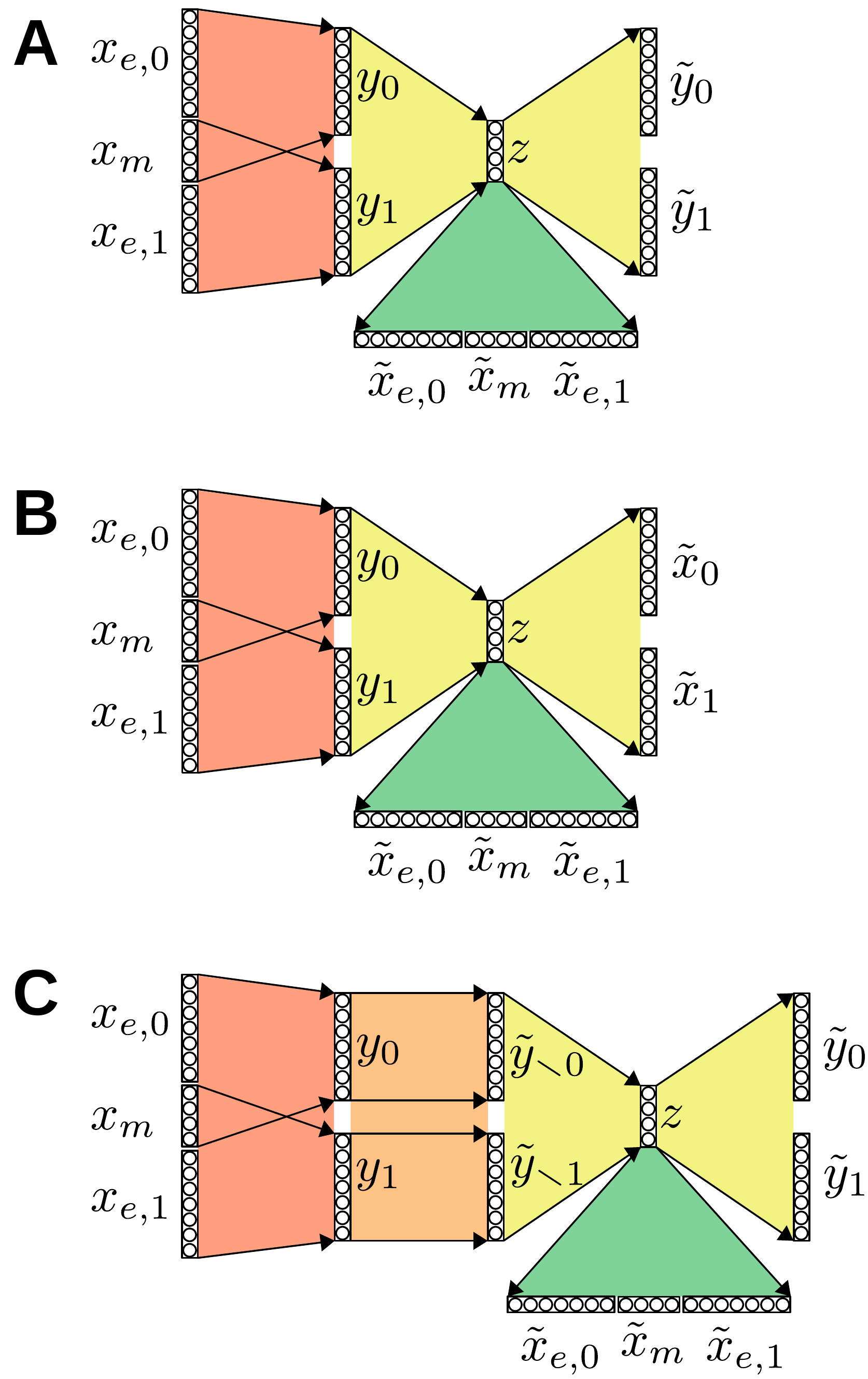}
    \caption{Overview of the approaches, assuming only $2$ modalities ($n = 2$). \textbf{A} baseline experiment, jointly encoding {\charlescolor (JE)} the dependent vectors $y_i$. \textbf{B} control experiment, jointly encoding the dependent vectors $y_i$ but reconstructing the original data $x_i$. \textbf{C} cross-modality prediction experiment {\charlescolor (CM)}, jointly encoding the predicted vectors $\tilde{y}_{\smallsetminus i}$. In each schema, the red areas represent the random neural networks generating dependent vectors (section \ref{methods:synthetic_data_generation}), the yellow areas represent the encoding and decoding networks (section \ref{methods:synthetic_encoding}), the green areas represent the readout networks (section \ref{methods:synthetic_readout}), and the orange area represents the cross-modality prediction networks (section \ref{methods:synthetic_cross_modality}).}
\label{fig-networks}
\end{figure}

\section{Methods}

{\revisioncolor 
Our general approach comprises three processing steps. The first step consists in generating multimodal sensory data. In order to have more control over the amount of mutual information among the different sensory modalities, we present a way to generate the latter from noise. We call this type of data {\em synthetic} as it has no real significance. For this we use random multi-layer neural networks that map independent information sources $x$ onto different ``views'' $y$ seen by different sensory modalities. This models the process how multimodal sensory information is generated from unobserved causes in the world. Our approach applied to the synthetic data will be explained in sub-section \ref{methods:synthetic}. We also test our approach on multimodal data from an actual {\em robot} simulation. This setup will be introduced in sub-section \ref{methods:real}.
}

In the second step, we train a neural network autoencoder with varying capacity, i.e., size of the central bottleneck, to learn a compressed (lossy) representation $z$ from the concatenation of the multimodal inputs $y$. This models the process of a developing agent learning an abstract representation from multimodal sensory information. In the third step we analyze the learned representation $z$ and measure how much information it retains that is unique to individual modalities versus shared among multiple modalities. For this we train a third set of neural networks to reconstruct the original information sources $x$ from the latent code $z$. The reconstruction error is a proxy for how much information has been lost during the encoding process. We now explain the three processing steps in detail {\revisioncolor starting with the synthetic setup}.

\subsection{Experiments with Synthetic Multimodal Input}
\label{methods:synthetic}
\subsubsection{Step 1: Generating Synthetic Multimodal Input}
\label{methods:synthetic_data_generation}

We produce the multimodal sensory data in such a way that we can precisely control the amount of mutual information between the different sensory modalities. {\revisioncolor We first define a distribution $p_\mutual$ from which we sample information that is shared by all sensory modalities. $p_\mutual$ therefore represents independent information sources in the world that affect multiple sensory modalities.} We define it as a $d_m$-dimensional distribution of independent standard normal distributions. We then define a second distribution $p_\exclusive$, from which the information exclusive to each modality is sampled. We define $p_\exclusive$ as a $d_\exclusive$-dimensional distribution of independent standard normal Gaussians. For a shared vector $x_\mutual \sim p_\mutual$ and $n$ vectors $x_{\exclusive,i} \sim p_\exclusive$, $i \in \mathbb{N}_{n-1}$ we can create the vectors $x_i = x_{\exclusive,i} \oplus x_\mutual$ carrying the information of each modality, where $\oplus$ is the concatenation operation.

Our sensory modalities do not sample the underlying causes directly (e.g., objects and light sources), but indirectly (e.g., images provided by the eyes). To mimic such an indirect sampling of the world without making any strong assumptions, we generate the sensory inputs $y$ that the learning agent perceives via random neural networks. Specifically, we define the input to modality $i$ as:

\begin{equation}
    y_i = \frac{C\left(x_i, \theta_{C,i}\right) - \mu_i}{\sigma_i} \; , \label{correlation_network}
\end{equation}

where $C$ and $\theta_{C,i}$ are the input construction network and its weights for the modality $i$ and $\mu_i$ and $\sigma_i$ are constants calculated to normalize the components of $y_i$ to zero mean and unit variance.

Tuning the amount of mutual information between the vectors $y_i$ is done by changing the dimensionalities $d_\mutual$ and $d_\exclusive$ of the vectors $x_\mutual$ and $x_{\exclusive,i}$, respectively. The amount of information preserved from the vectors $x_i$ in the vectors $y_i$ depends on the dimension $d_y$ of the vectors $y_i$. We define $d_y$ to be proportional to the dimension $d_x = d_\mutual + d_\exclusive$:

\begin{equation}
    d_y = k \times d_x \; ,
\end{equation}
where $k \gg 1$. This ensures that the sensory inputs $y_i$ essentially retain all information from the sources $x_\mutual$ and $x_{\exclusive,i}$.

\subsubsection{Step 2: Learning an Abstract Representation of the Synthetic Multimodal Input via Autoencoding}
\label{methods:synthetic_encoding}

Taken together, the set of vectors $\{y_i\}_{i \in \mathbb{N}}$ carries once the information from each $x_{\exclusive,i}$ and $n$ times the information from the mutual vector $x_\mutual$. To show that a lossy-compression algorithm achieves a better encoding when favoring the reconstruction of the repeated information, we train an autoencoder to jointly encode the set of the $y_i$. We therefore construct the concatenation $y = y_0 \oplus \dots \oplus y_{n-1}$ to train the autoencoder:

\begin{align}
    z &= E\left(y, \theta_E\right) \\
    \tilde{y} &= D\left(z, \theta_D\right) \; ,
\end{align}
where $E$ and $\theta_E$ are the encoding network and its weights and $D$ and $\theta_D$ are the decoding network and its weights. Tuning the dimension $d_z$ of the latent representation $z$ enables us to control the amount of information lost in the encoding process. The training loss for the weights $\theta_E$ and $\theta_D$ is the mean squared error between the data and its reconstruction, averaged over the component dimension and summed over the batch dimension:

\begin{align}
    L_{E,D} &= \sum_\text{batch} \frac{1}{n d_y} \left(y - \tilde{y}\right)^2 \; . \label{autoencoder_loss}
\end{align}


\subsubsection{Step 3: Quantifying Independent and Shared Information in the Learned Latent Representation}
\label{methods:synthetic_readout}

Finally, in order to measure what information is preserved in the encoding $z$, we train readout neural networks to reconstruct the original data $x_\mutual$ and the vectors $x_{\exclusive,i}$:

\begin{align}
    \tilde{x}_\mutual &= R_\mutual\left(z, \theta_\mutual\right) \\
    \tilde{x}_{\exclusive,i} &= R_\exclusive\left(z, \theta_{R,i}\right) \text{,}
\end{align}
where $R_\mutual$ and $\theta_\mutual$ are the mutual information readout network and its weights, and the $R_\exclusive$ and $\theta_{R,i}$ are the \textit{exclusive information} readout networks and their weights.

The losses for training the readout operations are the mean squared errors between the readout and the original data summed over the batch dimension and averaged over the component dimension:

\begin{align} 
    L_\mutual &= \frac{1}{d_\mutual}\sum_\text{batch} \left(\tilde{x}_\mutual - x_\mutual\right)^2 \; \text{and}\\
    L_{\exclusive,i} &= \frac{1}{d_\exclusive}\sum_\text{batch} \left(\tilde{x}_{\exclusive,i} - x_{\exclusive,i}\right)^2 \; \text{.}
\end{align}

Finally, once the readout networks trained, we measure the average \textit{per data-point} mean squared errors

\begin{align}
    r_\mutual &= \frac{1}{d_\mutual} \mathbb{E}\left[\left(\tilde{x}_\mutual - x_\mutual\right)^2\right] \; \text{and}\\
    r_\exclusive &= \frac{1}{d_\exclusive} \mathbb{E}\left[\left(\tilde{x}_{\exclusive,i} - x_{\exclusive,i}\right)^2 \right] \; \text{,}
\end{align}
serving as a measure of the portion of the mutual and exclusive data retained in the encoding $z$.

As a control condition, we also study a second encoding mechanism, where, instead of reconstructing the dependent vectors $y_0 \oplus \dots \oplus y_{n-1} = y$, the decoder part reconstructs the original data $x_0 \oplus \dots \oplus x_{n-1} = x$. The loss for training the encoder and decoder networks $E$ and $D$ from equation \ref{autoencoder_loss} then becomes:

\begin{equation}
    L_{E,D} = \sum_\text{batch} \frac{1}{n d_x} \left(x - \tilde{x}\right)^2 \; \text{.}
\end{equation}

\subsubsection{An Alternative to Step 2: Isolating the Shared Information}
\label{methods:synthetic_cross_modality}

We also compare the previous approach to an alternative architecture, designed specifically to isolate the information shared between the modalities. Let $\left(A, B\right)$ be a pair of random variables defined over the space $\mathcal{A} \times \mathcal{B}$ with unknown joint distribution $P\left(A, B\right)$ and marginal distributions $P\left(A\right)$ and $P\left(B\right)$.
Determining the mutual information between the variables $A$ and $B$ consists in finding either one of $P\left(A,B\right)$, $P\left(A|B\right)$ or $P\left(B|A\right)$. With no other assumptions, this process requires to sample many times from the joint distribution $P\left(A,B\right)$. We propose to make the strong assumption that the conditional probabilities are standard normal distributions with a fixed standard deviation $\sigma$

\begin{align}
    P\left(B=b|A=a\right) = \mathcal{N}\left(\mu\left(a\right), \sigma, b\right)
\end{align}

We can then try to approximate the function $\mu\left(a\right)$ with a neural network $M\left(a, \theta_M\right)$ maximizing the probability $P\left(B=b|A=a\right)$. $\mu\left(a\right)$ thus represents the most likely $b$ associated with $a$, under the standard normal assumption. Training the network is done by minimizing the mean squared error loss

\begin{align}
    L_M 
    &= -\mathbb{E}_{a,b \sim P\left(A,B\right)}\left[\log\left(\mathcal{N}\left(\mu, \sigma, b\right)\right)\right] \\
    &= \mathbb{E}_{a,b \sim P\left(A,B\right)}\left[\left(\mu - b\right)^2\right] \cdot K_1 + K_2
\end{align}

with $K_1$ and $K_2$ constants depending on $\sigma$.

%
%
%
%

More concretely and using the notation from the first architecture, we define for each modality $i$ a neural network $M\left(y_i, \theta_{M_i}\right)$ learning to predict all other modality vectors $y_j, j \neq i$. The loss for the weights $\theta_{M_i}$ is defined

\begin{equation}
    L_{M_i} = \sum_\text{batch} \frac{1}{\left(n - 1\right) d_y} \left(y_{\smallsetminus i} - \tilde{y}_{\smallsetminus i}\right)^2
\end{equation}

with

\begin{equation}
    y_{\smallsetminus i} = \bigoplus_{j \neq i} y_j
\end{equation}

the concatenation of all vectors $y_j$ for $j \neq i$ and $\tilde{y}_{\smallsetminus i}$ the output of the network. We then consider the concatenation of the $\tilde{y}_{\smallsetminus i}$ for all $i$ as a high-dimensional code of the shared information.
This code is then compressed using an autoencoder, similarly to the description in Section~\ref{methods:synthetic_encoding}.  We vary the dimension of the encoder's latent code. Finally, similarly to the first approach, we train readout networks from the compressed latent code to determine how mutual and exclusive information are preserved in the process.

Overall, this way of processing the data is analogous to the baseline experiment in that the cross modality prediction networks and the subsequent auto-encoder, when considered together, form a network that transforms the vectors $y_i$ into themselves. Together, these two components can thus be considered as an auto-encoder, subject to a cross-modality prediction constraint.

\subsubsection{Neural Network Training}
\label{methods:synthetic_neural_network}

In the following, we compare three architectures against each other (compare Fig.~\ref{fig-networks}):

\begin{itemize}
    \item {\charlescolor The baseline architecture (JE), simply auto-encoding the} vectors $y_i$ jointly (cf.\ Fig.~\ref{fig-networks}A).
    \item The control condition with a simpler encoding task, where the vectors $y_i$ are encoded into a latent code $z$, from which the decoder tries to reconstruct the original vectors $x_i$, from which the inputs $y_i$ were generated (cf.\ Fig.~\ref{fig-networks}B).
    \item {\charlescolor The alternative architecture (CM), where for each modality,} a neural network tries to predict all other modalities and then all resulting predictions are jointly encoded, similarly to the baseline architecture (cf.\ Fig.~\ref{fig-networks}C).
\end{itemize}

We will now describe the training procedure and implementation details. In order to show the nature of the information preferably preserved by the encoding process, we measure the quality of the readouts obtained as we vary the dimension of the latent vector $d_z$. To this end, for each dimension $d_z \in \left[1 ; d_{z,max}\right]$, we successively train the cross modality prediction networks (experiment C only), the autoencoder weights $\theta_E$ and $\theta_D$ and the readout weights $\theta_\mutual$ and $\theta_{R,i}$. Once training is completed, we measure the average mean squared error of the readouts $\tilde{x}_\mutual$ and $\tilde{x}_{\exclusive,i}$. 

We choose the distributions of the vectors $x_\mutual$ and $x_{\exclusive,i}$ to be multivariate standard normal with a zero mean and unit variance. Therefore, a random guess would score an average mean squared error of $1$. Each experiment is repeated $3$ times and results are averaged.




The neural networks for the input construction $C$, cross-modality prediction $M$, encoding $E$, decoding $D$, mutual readout $R_\mutual$, and exclusive readout $R_\exclusive$ all have three fully-connected layers. The two first {\revisioncolor layers} always have a fixed dimension of \defaultlayerdim{} and use a ReLU as non-linearity. The final layer is always linear, its dimension for each network is reported in Table~\ref{network_dimension}.


\begin{table}
    \centering
    \begin{tabular}{c|cccccc}
        Network & $M$ & $C$ & $E$ & $D$ & $R_\mutual$ & $R_\exclusive$ \\
        \hline
        ~ \\
        Dimension & $\left(n - 1\right) \times d_y$ & $d_y$ & $d_z$ & $n \times d_y$ & $d_\mutual$ & $d_x$ \\
        ~\\
    \end{tabular}
    \caption{Dimension of the last layer of each neural network}
    \label{network_dimension}
\end{table}

For each model architecture A, B, or C, we show the effect of varying the ratio between mutual and exclusive data and that of varying the number of modalities. The default experiment used $d_\exclusive = 4$, $d_\mutual = 4$, $n = 2$, $k = 10$. We then varied $d_\exclusive \in \{4, 10, 16, 22\}$ or $n \in \{2, 3, 4, 5\}$, keeping all other parameters fixed.

Each network is trained on $2500$ batches of data of size $128$ with a learning rate of $10^{-3}$ and using the Adam algorithm \cite{adam}.

{\revisioncolor 

\subsection{Experiments with Multimodal Input from a Robot Simulation}
\label{methods:real}
\subsubsection{Step 1: Generating Multimodal Input from the Robot Simulation}
\label{methods:real_data_generation}

\begin{figure}
    \centering
    \includegraphics[width=\linewidth]{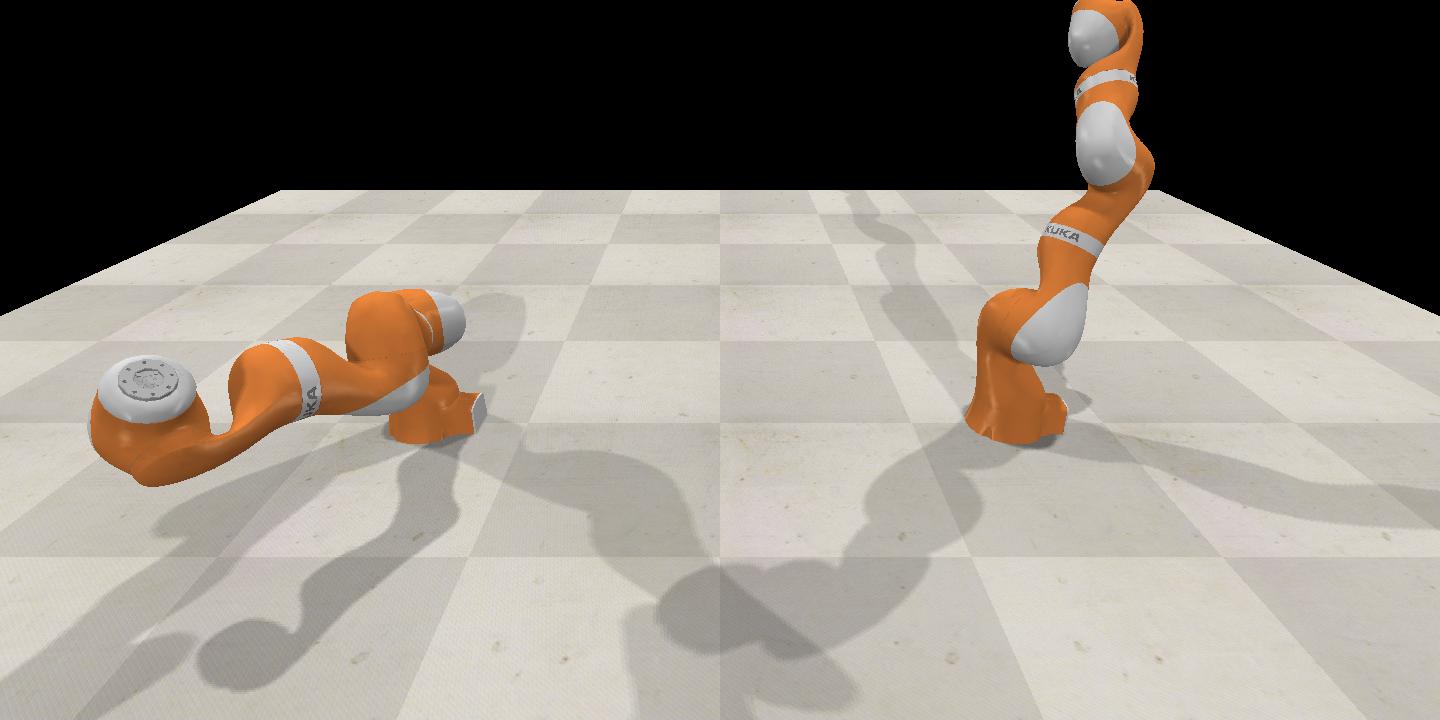}
    \caption{\revisioncolor High resolution image of the $2$ robot arms in the simulated environment. The images in the dataset have a resolution of $32$ by $64$ pixels only.}
    \label{fig:real_environment}
\end{figure}

In order to validate our approach in a more realistic setting, we applied it to data generated from a robot simulator, in which we placed two $7$-degrees-of-freedom robot arms side by side (see Fig.~\ref{fig:real_environment}). We then generated a dataset comprised of pictures of the $2$ arms, representing the visual modality, and of the joint positions and speeds, representing the proprioceptive modality. It has a total size of $2.000.000$ samples. To generate the dataset, we sampled random target joints positions uniformly in the joints' motion range, and we simulated $10$ iterations of $0.2$ seconds to let the agent reach the random target using its position PID controllers. At each iteration, a snapshots of the vision sensors and the joint sensors is recorded. Note how the position information is present both in the visual and proprioceptive modalities. However, as each data-sample is composed of a single image, the velocity information is present only in the proprioceptive modality. So as to also have at our disposal an information stream that is uniquely present in the visual modality, we decided to provide the encoding networks with only the proprioceptive information from one of the two arms (the right arm). Thus, the position information of the other arm (left arm) is only available through the visual information. Furthermore, by doing so, the velocity information about the left arm is present in neither of the two modalities and thus serves as a control factor. Finally, the dataset also contains records of the end effectors' positions of the two arms. We consider the end effector position of the right arm as being implicitly part of the proprioceptive modality, as it can be accurately deduced from the position information, while not directly feeding it into the networks. A summary of the information available to each modality is provided in Fig.~\ref{fig_modalities}. In Sec.~\ref{methods:synthetic_data_generation}, we named the vectors representing the different modalities $y_i$. When dealing with the realistic data, we will use $y_0 = y_\vision$ for the visual modality and $y_1 = y_\proprioception$ for the proprioceptive one. The $y_\proprioception$ is $z$-scored, i.e., it has a $0$ mean and a standard deviation of $1$. The $y_\vision$ vector is normalized such that the pixel values are in $\left[-1, 1\right]$.

We will now redefine the steps $2$, $3$ and the alternative to step $2$ for this dataset. The main difference with the synthetic dataset lies in the fact that the visual information is processed with convolutional neural networks. Moreover, we propose to compare $2$ ways of jointly encoding the modalities, which we refer to as  \textit{options}. The {\charlescolor default option} is analogous to the way the synthetic data is encoded, with the difference that the visual information is processed by a convolutional neural network. The second option {\charlescolor named Alternative Encoding Scheme (AES)} consists {\charlescolor in} learning a latent representation of the visual information with a convolutional autoencoder prior to jointly encoding the latent visual code with the proprioceptive information. 

\begin{figure}
    \centering
    \includegraphics[trim=0.0cm 0.0cm 0.0cm 7cm, width=\linewidth]{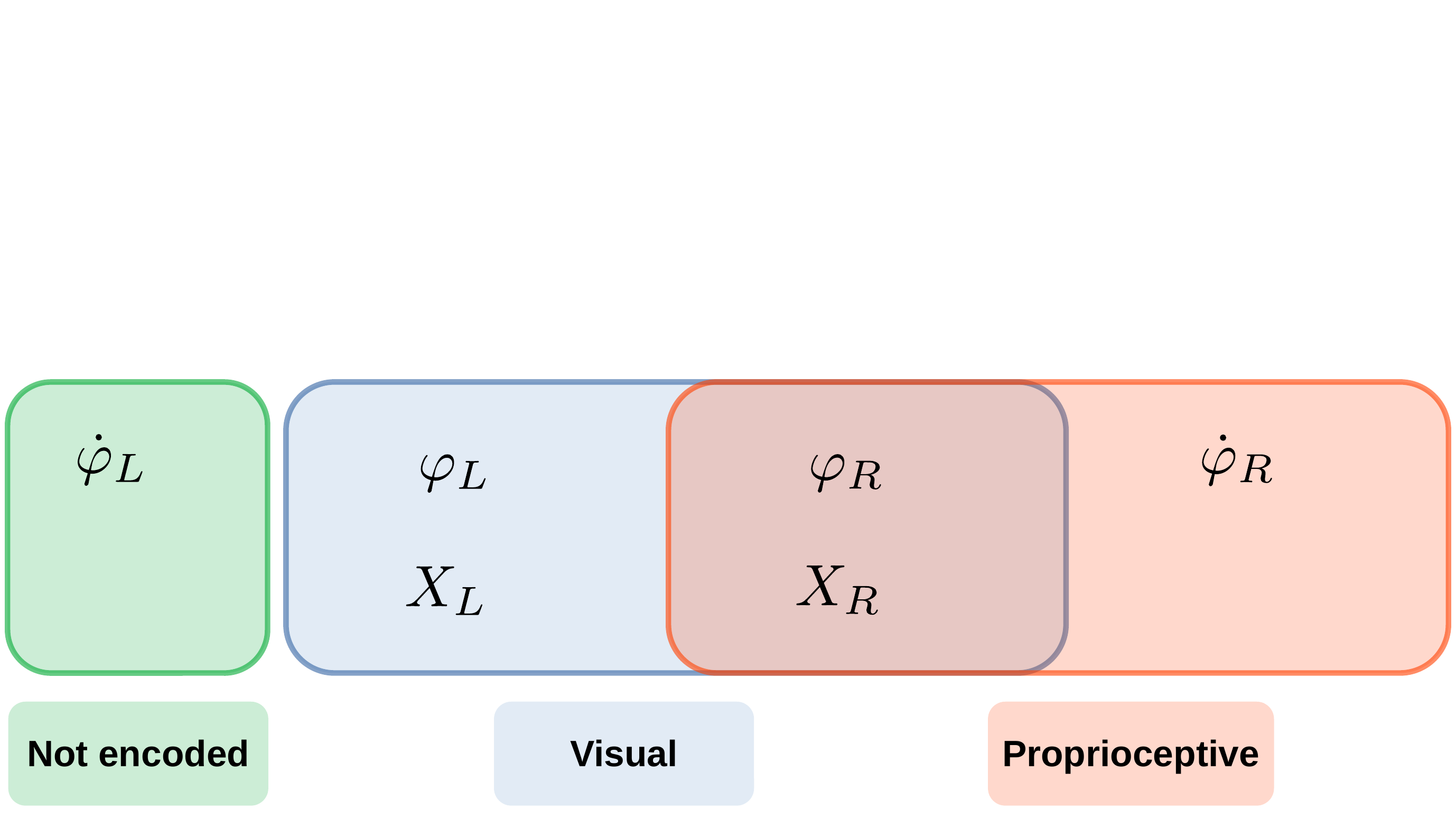}
    \caption{{\revisioncolor Schema representing the information available to each modalities for the realistic data dataset. $\varphi$ and ${\dot \varphi}$ denote the positions and velocities of the joints, respectively, and $X_L$ and $X_R$ the left / right parts of the visual information. Note that the positions and velocities of the left arm are not part of the proprioceptive modality. This way, the information about the position of the left arm is available only through the vision sensor. It also results that the velocity information of the left arm is present in neither of the two modalities, and thus serves as a control factor in our experiments. }}
    \label{fig_modalities}
\end{figure}

\subsubsection{Step 2: Learning an Abstract Representation of the Robot Data}
\label{methods:real_encoding}

Similar to Sec.~\ref{methods:synthetic_encoding}, the $y_\vision$ and $y_\proprioception$ vectors are jointly encoded and decoded with an autoencoder $\left(E, \theta_E, D, \theta_D\right)$. This time, however, the encoding and decoding steps are divided into two parts:

\begin{align}
    z_\preencoding &= E_\vision\left(y_\vision, \theta_{E_\vision}\right) \oplus E_\proprioception\left(y_\proprioception, \theta_{E_\proprioception}\right) \\
    z &= E_\preencoding\left(z_\preencoding, \theta_{E_\preencoding}\right)
\end{align}
for the encoder and
\begin{align}
    z_\postencoding &= D_\postencoding\left(z, \theta_{D_\postencoding}\right)\\
    \tilde{y}_\vision &= D_\vision\left(z_{\postencoding,\vision}, \theta_{D_\vision}\right)\\
    \tilde{y}_\proprioception &= D_\proprioception\left(z_{\postencoding,\proprioception}, \theta_{D_\proprioception}\right)
\end{align}
for the decoder, where $z_{\postencoding,\vision}$ and $z_{\postencoding,\proprioception}$ form a partition of $z_\postencoding$. The index within $z_\postencoding$ at which the split occurs is a hyper-parameter. The loss is then defined as:
\begin{align}
    L_{E,D} = \frac{1}{2 d_\proprioception} \sum_\text{batch} \left(\tilde{y}_\proprioception - y_\proprioception\right)^2 + \frac{1}{2 d_\vision} \sum_\text{batch} \left(\tilde{y}_\vision - y_\vision\right)^2 \; ,
\end{align}
where $d_\proprioception$ and $d_\vision$ are the sizes of the proprioception and vision tensors, respectively.

Dividing the encoding and decoding process in two parts enables to use convolutional and deconvolutional networks $E_\vision$ and $D_\vision$ to encode and decode the visual information. We did not find any difference when pre-encoding the proprioceptive information with a MLP compared to directly feeding it to the $E_\preencoding$ network, we will therefore report the results for $E_\proprioception = \text{id}$ and $D_\proprioception = \text{id}$.

{\charlescolor In the case of the Alternative Encoding Scheme (AES) option,} the $y_\vision$ is a compressed representation of the visual information and there is no need to process it with a convolutional neural network. In that case, we also set $E_\vision = \text{id}$ and $D_\vision = \text{id}$, meaning that the architecture of the networks in that case is the same as for the synthetic dataset.


\subsubsection{Step 3: Deciphering the Latent Code}
\label{methods:real_readout}

Similarly to Sec.~\ref{methods:synthetic_readout}, we train readout neural networks to decipher the information contained in the encoding $z$. This time, however, since we do not have access to the original vectors $x$ which induced the vectors $y_\vision$ and $y_\proprioception$, the readout operation aims at reconstructing the proprioceptive information from both arms $y_\target = y_{\position,\ensuremath{l}} \oplus y_{\velocity,\ensuremath{l}} \oplus y_{\eeff,\ensuremath{l}} \oplus y_{\position,\ensuremath{r}} \oplus y_{\velocity,\ensuremath{r}} \oplus y_{\eeff,\ensuremath{r}}$. The readout operation is written as:
\begin{align}
    y_\readout = R\left(z, \theta_\readout\right)
\end{align} and its loss is:
\begin{align}
    L_\readout = \frac{1}{d_\target} \sum_{\text{batch}} \left(y_\readout - y_\target\right)^2 \; .
\end{align}

\subsubsection{An Alternative to Step 2 for the Robot Data: Isolating the Shared Information}
\label{methods:real_cross_modality}

Finally, similarly to Sec.~\ref{methods:synthetic_cross_modality}, we propose an alternative to step number $2$ aiming at isolating only the information shared by both modalities. This is done by training two \textit{cross-modality prediction} networks:
\begin{align}
    \tilde{y}_{\smallsetminus\proprioception} &= M_\vision\left(y_\proprioception, \theta_{M_\vision}\right) \quad\text{and}\\
    \tilde{y}_{\smallsetminus\vision} &= M_\proprioception\left(y_\vision, \theta_{M_\proprioception}\right)
\end{align}
with the losses
\begin{align}
    L_{M_\vision} &= \sum_\text{batch}\frac{1}{d_\vision} \left(\tilde{y}_{\smallsetminus\proprioception} - y_\vision\right)^2 \quad\text{and}\\
    L_{M_\proprioception} &= \sum_\text{batch}\frac{1}{d_\proprioception} \left(\tilde{y}_{\smallsetminus\vision} - y_\proprioception\right)^2 \; .
\end{align}
Again, like in Sec.~\ref{methods:real_encoding}, the representations $\tilde{y}_{\smallsetminus\proprioception}$ and $\tilde{y}_{\smallsetminus\vision}$ are encoded using the autoencoder networks $E_\vision$, $E_\proprioception$, $E_\preencoding$, $D_\postencoding$, $D_\vision$ and $D_\proprioception$:
\begin{align}
    z_\preencoding &= E_\vision\left(\tilde{y}_{\smallsetminus\proprioception}, \theta_{E_\vision}\right) \oplus E_\proprioception\left(\tilde{y}_{\smallsetminus\vision}, \theta_{E_\proprioception}\right) \\
    z &= E_\preencoding\left(z_\preencoding, \theta_{E_\preencoding}\right) \\
    z_\postencoding &= D_\postencoding\left(z, \theta_{D_\postencoding}\right)\\
    \tilde{y}_\vision &= D_\vision\left(z_{\postencoding,\vision}, \theta_{D_\vision}\right)\\
    \tilde{y}_\proprioception &= D_\proprioception\left(z_{\postencoding,\proprioception}, \theta_{D_\proprioception}\right) \; .
\end{align}
For the AES option, since $y_\vision$ is a one-dimensional vector, we set $E_\vision = D_\vision = E_\proprioception = D_\proprioception = \text{id}$.

\subsubsection{Description of the Networks}
In the case of the default option, the network $E_\vision$ is a convolutional neural network composed of $2$ convolutional layers with kernel size $4$ and stride $2$ followed by a dense layer with output size $100$. The network $E_\preencoding$ is a $3$-layered MLP where all layer sizes but the last are $200$. The last layer uses a linear activation function and has a size $d_z$. The network $D_\postencoding$ is a $3$-layered MLP where all layer sizes but the last are $200$. The last layer uses a linear activation function and has a size $100 + d_\proprioception$. The network $D_\vision$ is a deconvolutional neural network composed of a dense layer of size $8192$ followed by two transposed convolutional layers with kernel sizes $4$ and strides $2$. Finally, the readout network is also a $3$-layered MLP where all layer sizes but the last are $200$. The last layer uses a linear activation function and has a size $d_\target$.

For the cross-modality prediction, the network $M_\vision$ is a deconvolutional neural network composed of a dense layer of size $8192$ followed by two transposed convolutional layers with kernel size $4$ and stride $2$ and the network $M_\proprioception$ is a convolutional neural network composed of two convolutional layers with kernel size $4$ and stride $2$ followed by a dense layer of size $d_\proprioception$.

Finally, as stated above, in the case of the AES option, $y_\vision$ is a learned code of size $100$ representing the visual information. In this case we set $E_\vision = D_\vision = E_\proprioception = D_\proprioception = \text{id}$. The network learning the code is a convolutional autoencoder composed of $2$ convolutions, one dense layer of size $100$, one dense layer of size $8192$ and $2$ transposed convolutions.
}

\section{Results}

\subsection{Lossy Compression of Multimodal Input Preferentially Encodes Information Shared Across Modalities}
\label{results:synthetic_shared_prefered}

\begin{figure}
    \centering
    \includegraphics[width=\linewidth]{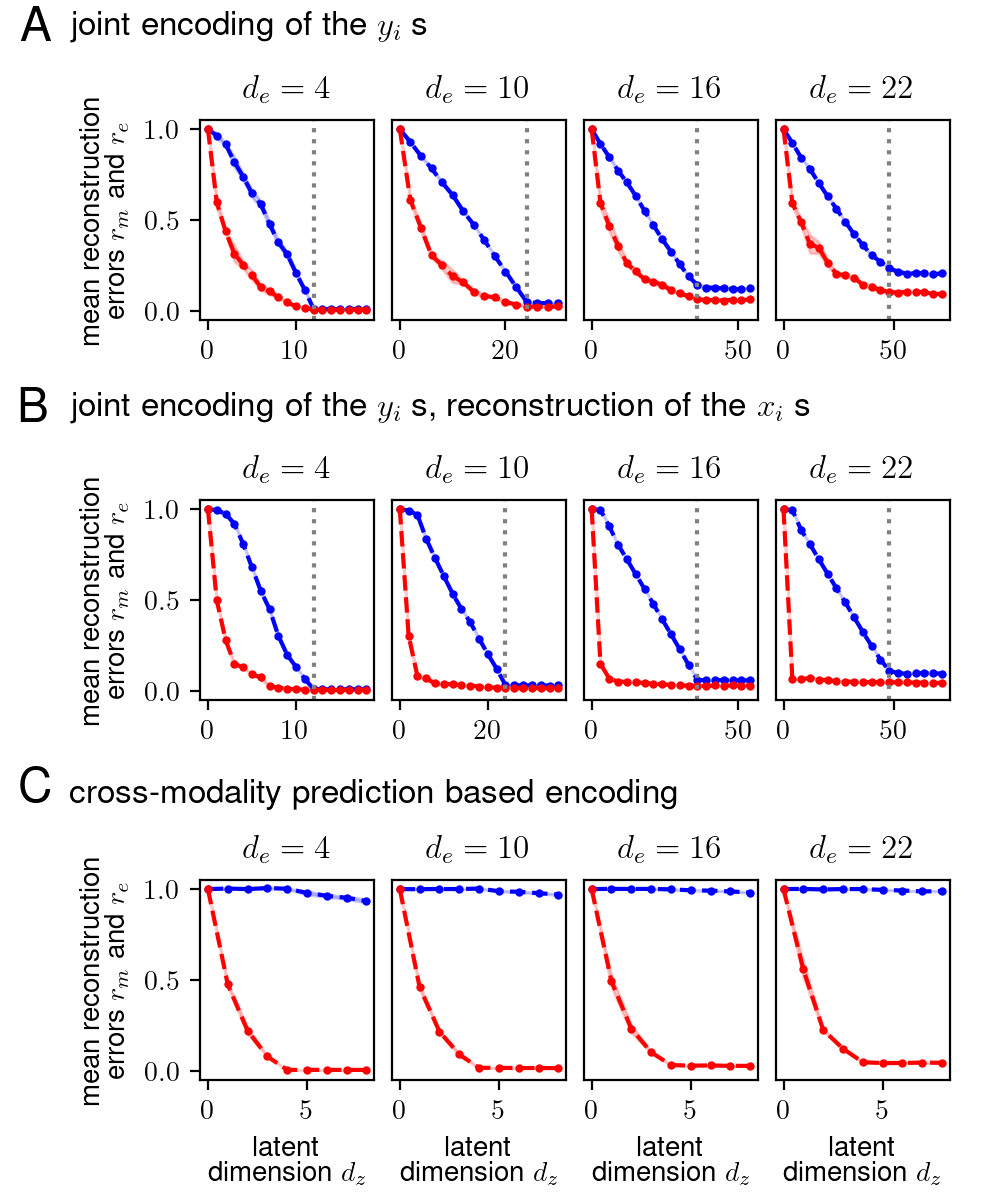}
    \caption{Each plot represents the reconstruction error of the readout operation for the exclusive data $r_\exclusive$ in blue, and for the shared data $r_\mutual$ in red, as a function of the auto-encoder latent dimension. The dotted vertical line indicates the latent dimension matching $n d_\exclusive + d_\mutual$. The data point for a latent dimension of $0$ is theoretically inferred to be equal to $1.0$ (random guess). The four plots in one row correspond to different dimensions $d_\exclusive$ of the exclusive data. The results are presented for the three architectures \textbf{A}, \textbf{B} and \textbf{C}.}
    \label{fig:vary_exclusive_dim}
\end{figure}

Figure~\ref{fig:vary_exclusive_dim} shows the reconstruction errors for exclusive vs.\ shared information as a function of $d_z$, the size of the autencoder's bottleneck, for the three different architectures. Each data point represents the mean of $3$ repetitions of the experiments, and the shaded region around the curves indicate the standard deviation.

The grey dotted vertical line indicates the latent code dimension $d_z$ matching the number of different univariate gaussian distributions used for generating the correlated vectors $y_i$, $d_{min} = d_\mutual + n d_\exclusive$. Assuming that each dimension in the latent vector can encode the information in one normally distributed data source, when $d_z = d_{min}$ both the exclusive data and the shared data can theoretically be encoded with minimal information loss. Knowing that random guesses would score a reconstruction error of $1.0$, we can augment the data with the theoretical values $r_\mutual = 1$ and $r_\exclusive = 1$ for $d_z = 0$.

The results for {\charlescolor the JE architecture (cf. Fig.~\ref{fig-networks}A)}, jointly encoding the correlated vectors $y_i$, show that the data shared by all modalities is consistently better reconstructed by the autoencoder for all latent code sizes $d_z$. In particular, this is also true for over-complete codes when $d_z > d_{min}$. Information loss in that regime is due to imperfections of the function approximator. When the code dimension is bellow $d_{min}$, the information loss is greater, as not all of the data can pass through the antoencoder's bottleneck. These results confirm our intuition from the Introduction that shared information should be preferentially encoded during lossy compression of multimodal inputs.

This information filtering is a consequence of neural networks' continuity, implying topological properties on the functions that they can learn. Indeed, while there exist non-continuous functions for which the dimensionality of the codomain is greater than that of the domain, the continuity property enforces that the dimension of the codomain is less or equal to that of the domain. As a consequence, the dimensionality of the codomain of the decoder network of an autoencoder is less than or equal to the dimensionality of the latent code. If the dimension of the latent code is itself lower than that of the data, as can be enforced by a bottleneck, it follows that the data and its reconstruction sit on manifolds of different dimensionality, implying information loss.


In the \textit{under-complete} regime, $d_z < d_{min}$, the autoencoder shows a stronger preference for retaining the shared data, partly filtering out the exclusive data. The chief reason for this is that the shared data is essentially counted $n$ times in the network's reconstruction loss, while the exclusive data is counted only once. As the dimension of the exclusive data $d_\exclusive$ increases, we still observe the two training regimes for $d_z$ less or greater than $d_{min}$, even though the boundary between both tends to vanish as we reach the network's capacity.

The results for the second (control) architecture {\charlescolor (cf. Fig.~\ref{fig-networks}B)}, jointly encoding the correlated vectors $y_i$ by reconstructing the original data vectors $x_i$ rather than $y_i$, are {\charlescolor similar in nature to those of the JE experiment}. The main differences occur at low values of $d_z$. The readout quality of the exclusive data is overall higher and that of the shared data lower. 

Finally, results for {\charlescolor the CM architecture (cf. Fig.~\ref{fig-networks}C)}, encoding the cross-modality predictions, are significantly different and confirm that it is possible to isolate the mutual information between different data sources. Notice how for $d_\exclusive = 4$, the readout quality of the exclusive data $r_\exclusive$ seems to improve slowly as the dimension $d_z$ increases. We verified that the values of $r_\exclusive$ remain high for high values of $d_z$, measuring reconstruction errors converging around $0.8$. Thus, this architecture is more effective in stripping away any exclusive information. This is because, by definition, exclusive information cannot be encoded during the initial cross-modality prediction (Fig.~\ref{fig-networks}C, orange part).

\subsection{Increasing the Number of Modalities Promotes Retention of Shared Information}
\label{results:synthetic_repeated_prefered}

\begin{figure}
    \centering
    \includegraphics[width=\linewidth]{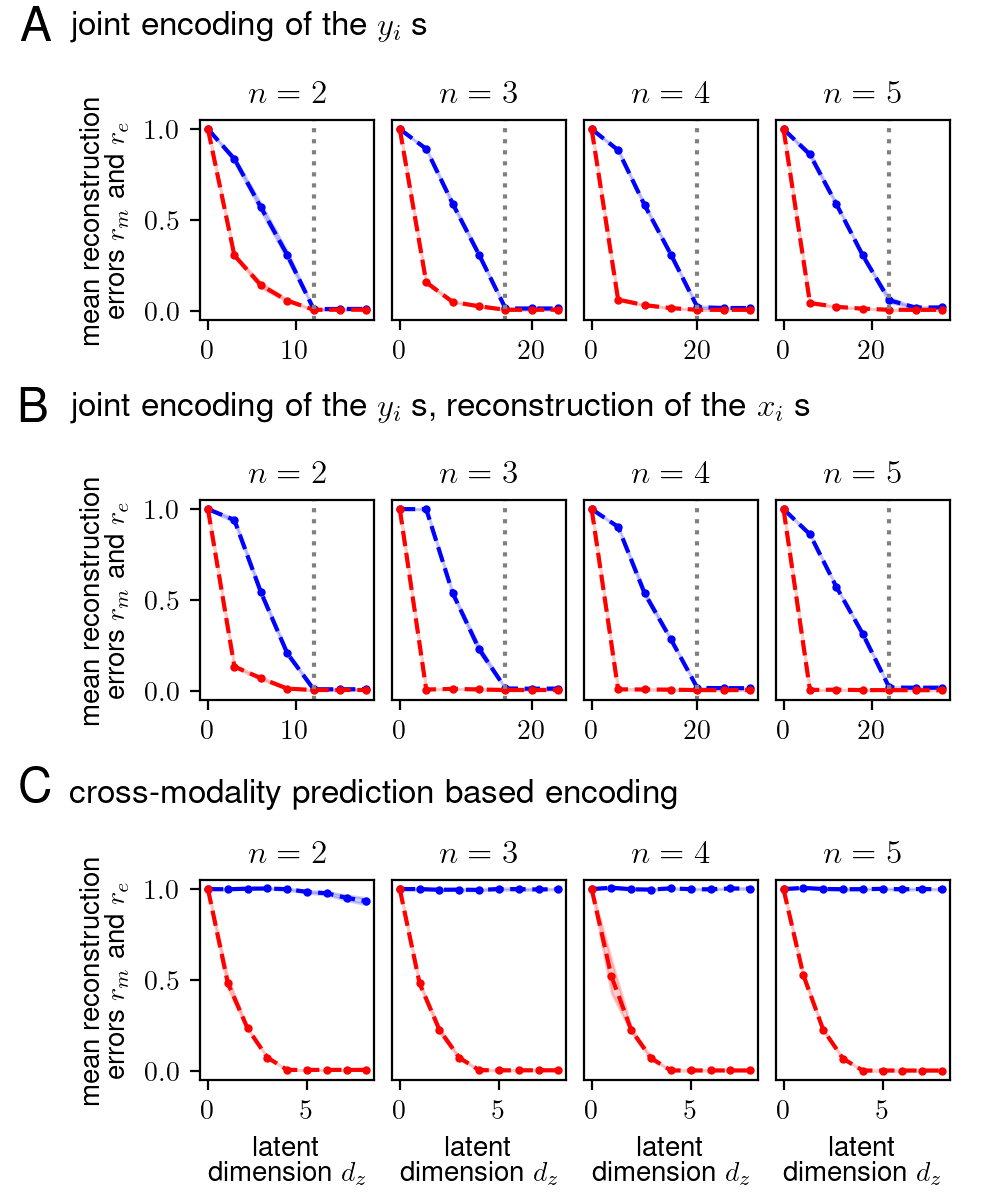}
    \caption{Similarly to Fig.~\ref{fig:vary_exclusive_dim}, each plot represents the reconstruction error of the readout operation for the exclusive data $r_\exclusive$ in blue, and for the shared data $r_\mutual$ in red, as a function of the auto-encoder latent dimension. The four plots correspond to a different number $n$ of modalities. The results are presented for the three architectures \textbf{A}, \textbf{B} and \textbf{C}.
    }
    \label{fig:vary_n_sources}
\end{figure}

Figure~\ref{fig:vary_n_sources} shows the results for varying the number of modalities. {\charlescolor For the JE and control architectures, they show how} increasing the number of modalities reinforces the retention of the shared data over the exclusive data. Note how the reconstruction errors for the shared information (red curves) decay more rapidly for higher numbers of modalities $n$. This is {\charlescolor in contrast to the CM architecture}, where results are very similar for different numbers of modalities. This is because the initial cross-modality prediction network (Fig.~\ref{fig-networks}C, orange part) effectively removes all modality-specific information, leaving essentially the same encoding task for the subsequent autoencoder despite the different numbers of modalities $n$.

{\revisioncolor 
\subsection{Results on the Robot Dataset}
\label{results:real}

\begin{figure}
    \centering
    \includegraphics[width=\linewidth]{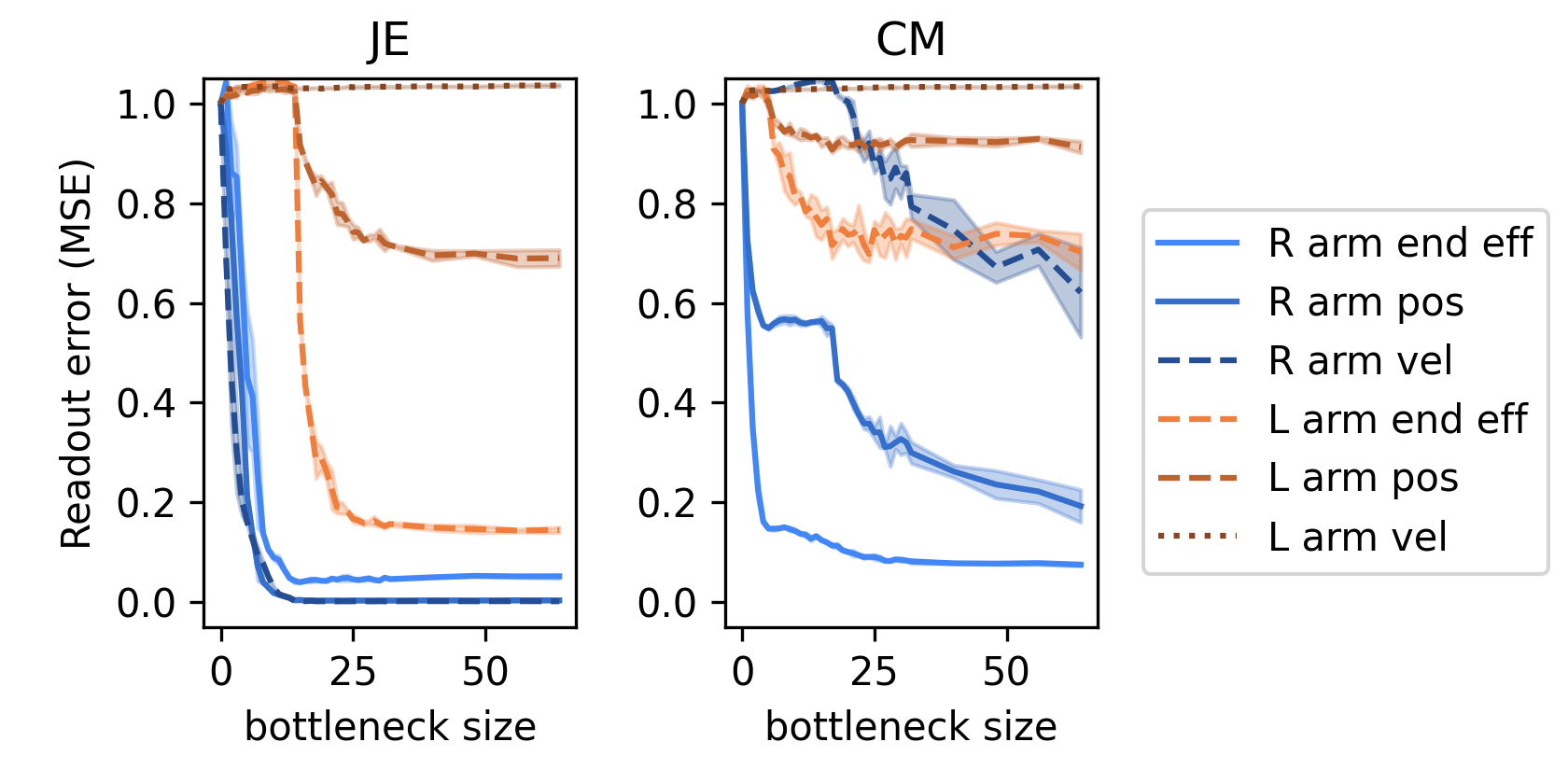}
    \caption{Readout reconstruction errors {\charlescolor for the JE and CM approaches} as a function of the size of the bottleneck of the encoding. Blue and red curves correspond to right and left arms respectively, solid lines correspond to information present in both modalities, dashed lines to information present in one modality only, and the dotted line to information present in none of the modalities.}
    \label{fig:real_readout_option_1}
\end{figure}

{\charlescolor Figure \ref{fig:real_readout_option_1} shows the readout errors for the proprioceptive information from both arms, for the JE and CM architecture, as a} function of the size of the latent code. In both cases, the velocity information about the left arm, which is present in neither of the $2$ modalities and thus serves as a control factor, is not recovered for any latent code size. This is indicated by a chance-level reconstruction quality of $1.0$. {\charlescolor For the JE architecture,} the information which is present in both modalities (i.e. the right arm's joint positions and the right end-effector position) is well reconstructed. {\charlescolor For the CM architecture,} however, the right arm's joint positions are recovered with a MSE of around at best $0.2$. The reason for this is that the position of some of the joints is not visible at all in the image frames (like for example the last joint in the arm rotating the wrist). Furthermore, in some positions occlusion effects occur. The information about these joints is therefore not present in the mutual information and is thus filtered out by the cross-modality prediction. The joint velocity information inherently has a low entropy, making it easier to compress. As a result, {\charlescolor the JE approach is very good at} recovering this information. The other approach, however, has properly filtered it out, even-though some of the information seems to have leaked out during the proprioception $\rightarrow$ vision cross-modality prediction. This is understandable given the big increase in dimensionality taking place in this operation. {\charlescolor For the CM approach,} the position of the left arm, which is present only in the visual modality, is properly filtered out with MSEs greater than $0.7$ for all latent sizes. In the first approach however, the left end-effector position is recovered only for latent sizes greater than $14$, which corresponds to the point where the proprioception of the right arm is fully represented in the latent code.

\begin{figure}
    \centering
    \includegraphics[width=\linewidth]{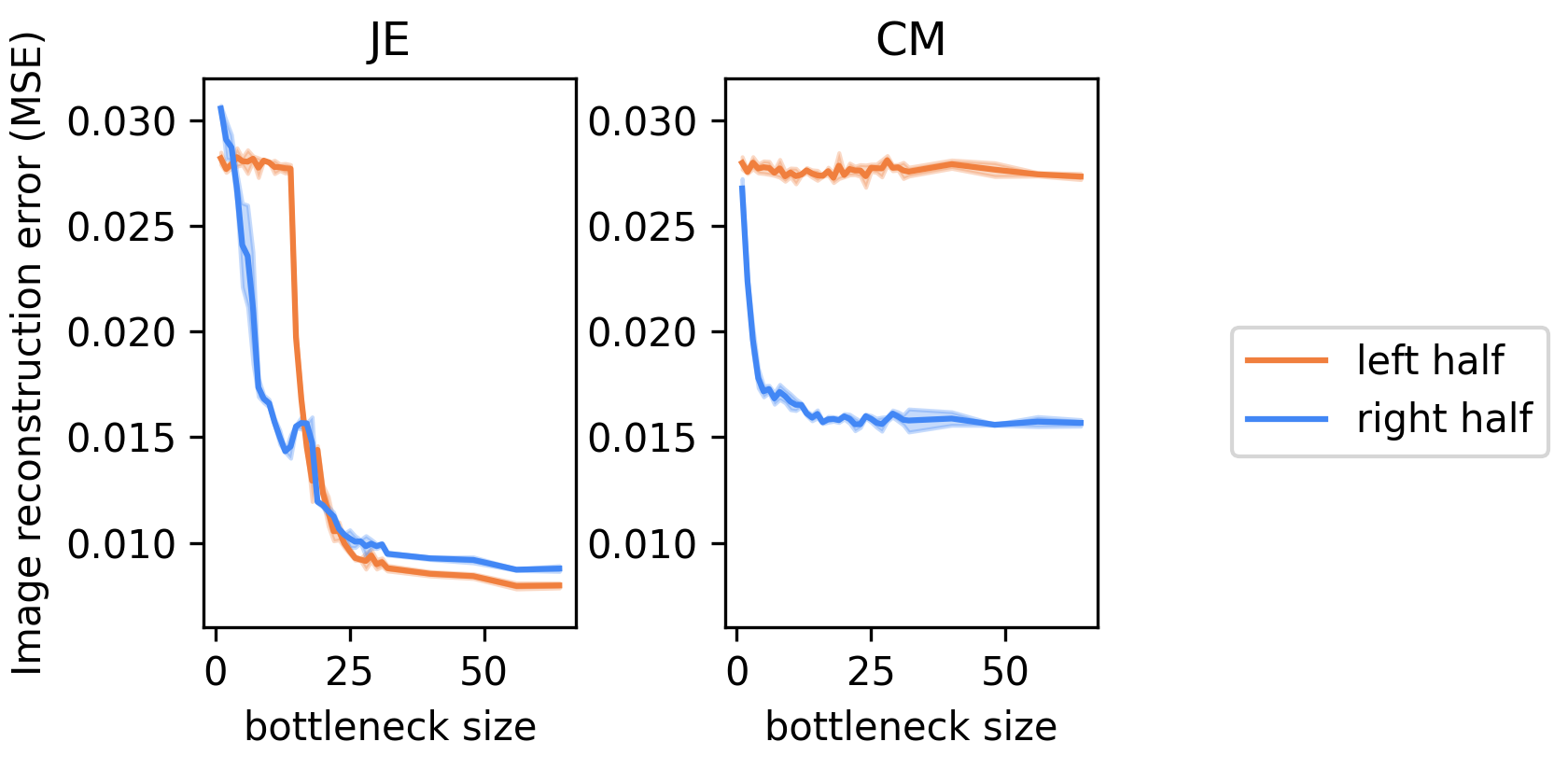}
    \caption{Reconstruction error of the visual modality {\charlescolor for the JE and CM approaches} as a function of the size of the bottleneck of the encoding. The error is split in two parts corresponding to the left and right halves of the frames. The results show that the pixels which share information with the proprioceptive modality are better reconstructed.
    }
    \label{fig:real_vision_left_right_option_1}
\end{figure}

Figure \ref{fig:real_vision_left_right_option_1} shows the reconstruction error of the visual information as a function of the latent code size {\charlescolor for the JE and CM approaches}. The reconstruction error is split into two parts corresponding to the left and right half of the frames. Note that the chance reconstruction error is around $0.027$ (MSE). The results show that in both approaches, the pixels corresponding to the right arm are better reconstructed. {\charlescolor In the JE approach,} when the latent code size allows it, the entirety of the frame is encoded while {\charlescolor for the CM approach,} the left half of the frame is never encoded.

\begin{figure}
    \centering
    \includegraphics[width=\linewidth]{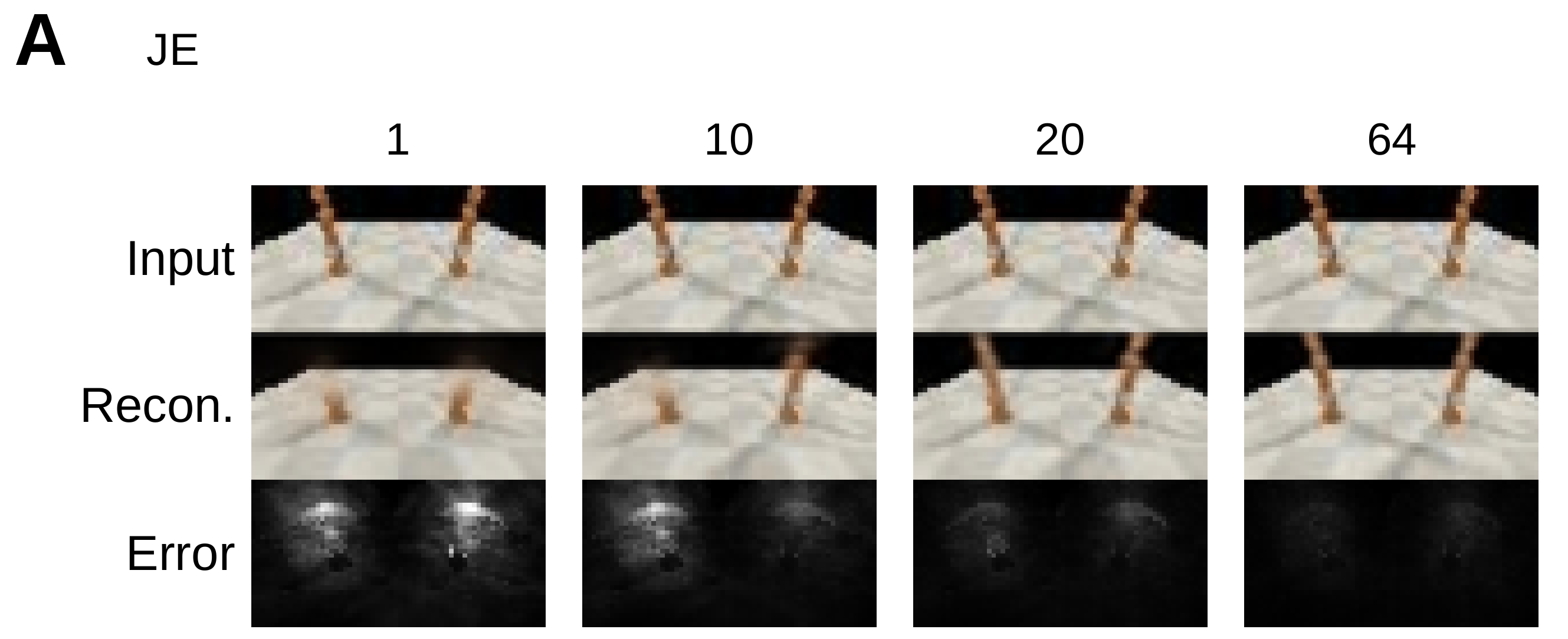}

    \vspace{0.6cm}

    \includegraphics[width=\linewidth]{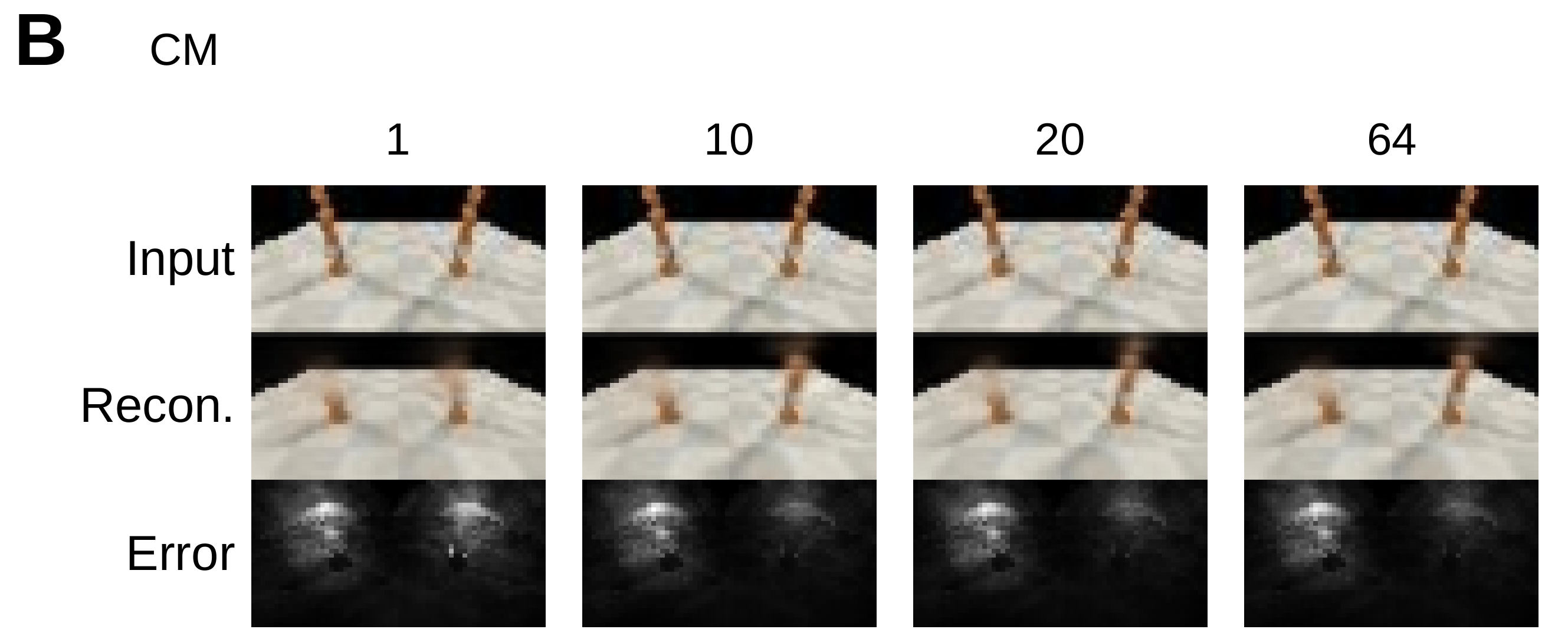}
    \caption{We show for each bottleneck sizes in $\{1, 10, 20, 64\}$ an image and it's reconstruction through the autoencoder, as well as the mean reconstruction error map (darker indicates lower error). {\charlescolor The JE approach (\textbf{A}) encodes the visual information about both arms if the information bottleneck allows for it, while the CM approach (\textbf{B})} reconstructs only one of the two arm for any bottleneck size.}
    \label{fig:real_vision_map_option_1}
\end{figure}

{\charlescolor Finally, Fig.~\ref{fig:real_vision_map_option_1} shows concrete reconstructions of images by the encoding process for bottleneck sizes taken from the set $\{1, 10, 20, 64\}$ for the JE and CM approaches. The results clearly show that the JM approach (\textbf{A}) goes from a regime where neither the left nor the right part of the frame is reconstructed, to a regime where the right part is, but not the left, to a regime where the whole frame is correctly reconstructed. For the CM approach (\textbf{B}), the system only goes through the $2$ first regimes.}
}

\section{Discussion}
{\revisioncolor 
Forming abstract representations is critical for higher-level intelligence. We have argued that the essence of abstraction is the lossy compression of information --- stripping away details to arrive at a representation that transcends the original context and more easily generalizes to new situations. In principle, this could be done in many different ways. The critical question is what information to strip away and what residual information to keep. Depending on the task, there may be different answers to this question. For example, in a supervised classification setting a system may learn to strip away any information that is irrelevant for determining the class label. Or in a reinforcement learning setting, an agent may attempt to strip away any information that is not helpful for predicting future rewards.

Here we have focused on an unsupervised approach for} the learning of abstract representations through lossy compression of multimodal information. Our key result is that lossy compression of multimodal inputs through autoencoding naturally favors the retention of information that is {\em shared} across the sensory modalities. Such shared information may be particularly useful for generalizing to new situations.

{\revisioncolor
We first demonstrated our approach using synthetic multimodal data and then validated it using a simulated embodied agent (two-armed robot). The results indicate that the approach scales well to a more realistic scenario with visual and proprioceptive information. However, to compensate for the vastly different dimensionality of visual and proprioceptive data, we used a more complex network architecture where visual information was first passed through several convolutional layers before being integrated with proprioceptive information.
}

{\revisioncolor 
It is important to stress that different sensory modalities have evolved in biological organisms (and are used in robots) exactly {\em because they provide different, complementary information} about the world. Discarding information that is not shared among modalities but unique to a single modality therefore seems to undermine this modality's raison d'\^{e}tre. Indeed, we are {\bf not} arguing that such information is not useful and should be discarded altogether. What we are arguing is that such information may be {\bf less} useful when the goal is to learn highly {\em abstract} and compressed representations of the physical world. One of the greatest challenges for a developing mind is to make sense of what William James called the ``blooming, buzzing, confusion'' of sensations provided by different modalities, which eventually must become ``coalesced together into one and the same space'' \cite{james1890principles}. The challenge thus is to identify how the inputs provided by the different sensory modalities relate to one another, i.e., what information they share. As we have seen, our generic approach is able to distill this shared information from raw sensor data.
}

In this work, we have focused on generic autoencoder networks, as they are popular tools for dimensionality reduction and learning compressed representations of sensory signals in many contexts. In deep reinforcement learning, for example, they are frequently used to learn a compact abstract representation of high-dimensional (e.g., visual) input. In the future, it will be interesting to consider extensions of the generic autoencoder framework such as sparse autoencoders \cite{ksparse, sparse} or other forms of regularized autoencoders such as beta-variational autoencoders \cite{beta_vae}, or encoder networks that learn to simultaneously predict rewards to focus limited encoding resources on relevant aspects of the multimodal sensory inputs that are associated with rewards.

Our approach to learning abstract representations from multimodal data can also be related to approaches that learn view-invariant visual object representations through temporal coherence \cite{Foldiak1991, einhauser}.  In such approaches, temporal input from a single modality, typically the visual one, is considered, and a code is learned that is slowly varying, for example through trace rules or slow feature analysis (SFA).
This corresponds to a lossy compression across time: fast changing information is discarded and slowly changing information is retained. For example, the fast changing information may be the pose of the object and the slowly changing information may be the object's identity. Both jointly determine the current image. By retaining information that is slowly varying, a viewpoint invariant representation of the object's identity can be learned, which can be used to recognize the object in different poses. This form of temporal compression of information is complementary to the multimodal compression we have considered here. In fact, the simple information theoretic argument from the Introduction applies in the same way if we replace the visual and haptic inputs $X_v$ and $X_h$ with, say, successive inputs  from a single sensory modality. Lossy compression of such data will naturally favor the retention of information that the successive inputs have in common, i.e., information that is helpful for predicting $X_{t+1}$ from $X_t$ or vice versa. In the future, it will be interesting to consider learning abstract representations by abstracting across time and sensory modalities in hierarchical {\revisioncolor cognitive architectures for agents that learn abstract dynamic models of their interactions with the world}.

Multimodal compression as discussed here may also be an effective driver of intrinsically motivated exploration in infants and robots, reviewed in \cite{BaldassarreMirolli2013}. Schmidhuber proposed using compression progress as an intrinsic motivation signal \cite{Schmidhuber2008}. In our own work, we have proposed the Active Efficient Coding (AEC) framework, that uses an intrinsic motivation for coding efficiency \cite{zhao2012, teuliere2014, eckmann2020, wilmot2020}. AEC is an extension of classic efficient coding ideas \cite{barlow1961} to active perception. Next to learning an efficient code for sensory signals, it proposes to control behavior in a way to maximize the information coming from different sources, while reducing the reconstruction loss during lossy compression. This has been shown to lead to the fully autonomous self-calibration of active perception systems, e.g., active stereo vision or active motion vision. However, this approach has not been considered in a multimodal setting. For example, consider an infant (or a developing robot) moving her hand in front of her face. In this situation, visual and proprioceptive signals are coupled. As the hand is felt to move to the left, the visual sense indicates motion to the left. Thus, the signals can be jointly encoded more compactly or with less reconstruction loss. An intrinsic motivation trying to minimize such reconstruction loss will therefore promote behaviors, where signals from the different modalities are strongly coupled. For example, banging a toy on the table creates correlated sensations in visual, proprioceptive, haptic, and auditory modalities and affords strong compression when jointly encoding these signals. We conjecture that AEC-like intrinsic motivations may drive infants to engage in such behaviors and may be effective in guiding open-ended learning {\revisioncolor in robots who try to {\em understand} the world around them}.


%

\appendices
\section{Futility of a Symmetric Information Bottleneck}



{\revisioncolor 
We start from the original Information Bottleneck objective:
\begin{equation}
    \min_{p(t|x)} I(X;T) - \beta I(T;Y),
\end{equation}
where $p(T=t|X=x)$ describes the encoding of an input $x$ via its representation $t$, $I(X;T)$ is the mutual information between $X$ and $T$, $I(T;Y)$ is the mutual information between $T$ and $Y$, and $\beta$ serves to balance the two objectives.

We now consider the reverse problem of $T$ trying to extract information from $Y$ that is useful for predicting $X$. This leads to the following ``reverse'' objective:
\begin{equation}
    \min_{p(t|y)} I(Y;T) - \beta I(T;X),
\end{equation}
where the roles of $X$ and $Y$ have simply been swapped.

A naive approach to derive a symmetric version of the information bottleneck is to consider an encoding where $T$ encodes $X$ and $Y$ jointly via a probability $p(t|x,y)$, while minimizíng both the forward and the reverse functionals: 
\begin{equation}
    \min_{p(t|x,y)} I(X;T) - \beta I(T;Y) + I(Y;T) - \beta I(T;X),
\end{equation}
which simplifies to:
\begin{equation}
    \min_{p(t|x,y)} (1-\beta) I(X;T) + (1-\beta) I(Y;T)
\end{equation}
because of the symmetry of the mutual information (e.g.\ $I(X;T)=I(T;X)$). Unfortunately, however, for $\beta < 1$ this amounts to minimizing the information that $T$ contains about $X$ and $Y$, while for $\beta > 1$ this amounts to maximizing the information that $T$ contains about $X$ and $Y$. In neither case will $T$ contain {\em only} the information that is shared between $X$ and $Y$.
}

{\charlescolor
\section{Results for the AES Option}
}
~
\begin{figure}[!htb]
    \centering
    \includegraphics[width=\linewidth]{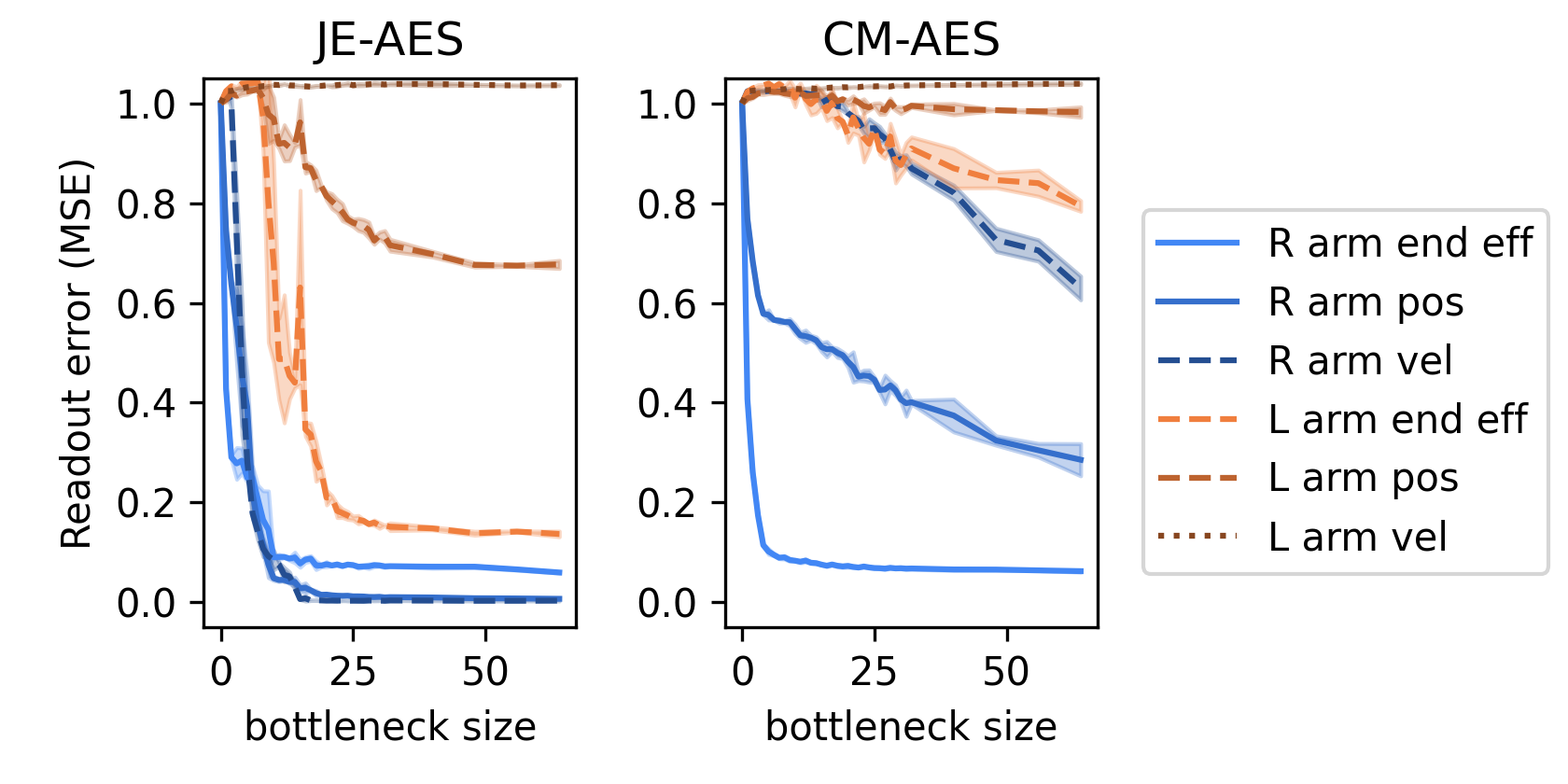}
    \caption{Readout reconstruction errors {\charlescolor for the JE and CM approaches} as a function of the size of the bottleneck of the encoding. Blue and red curves correspond to right and left arms respectively, solid lines correspond to information present in both modalities, dashed lines to information present in one modality only, and the dotted line to information present in none of the modalities.}
    \label{fig:real_readout_option_2}
\end{figure}

\begin{figure}[!htb]
    \centering
    \includegraphics[width=\linewidth]{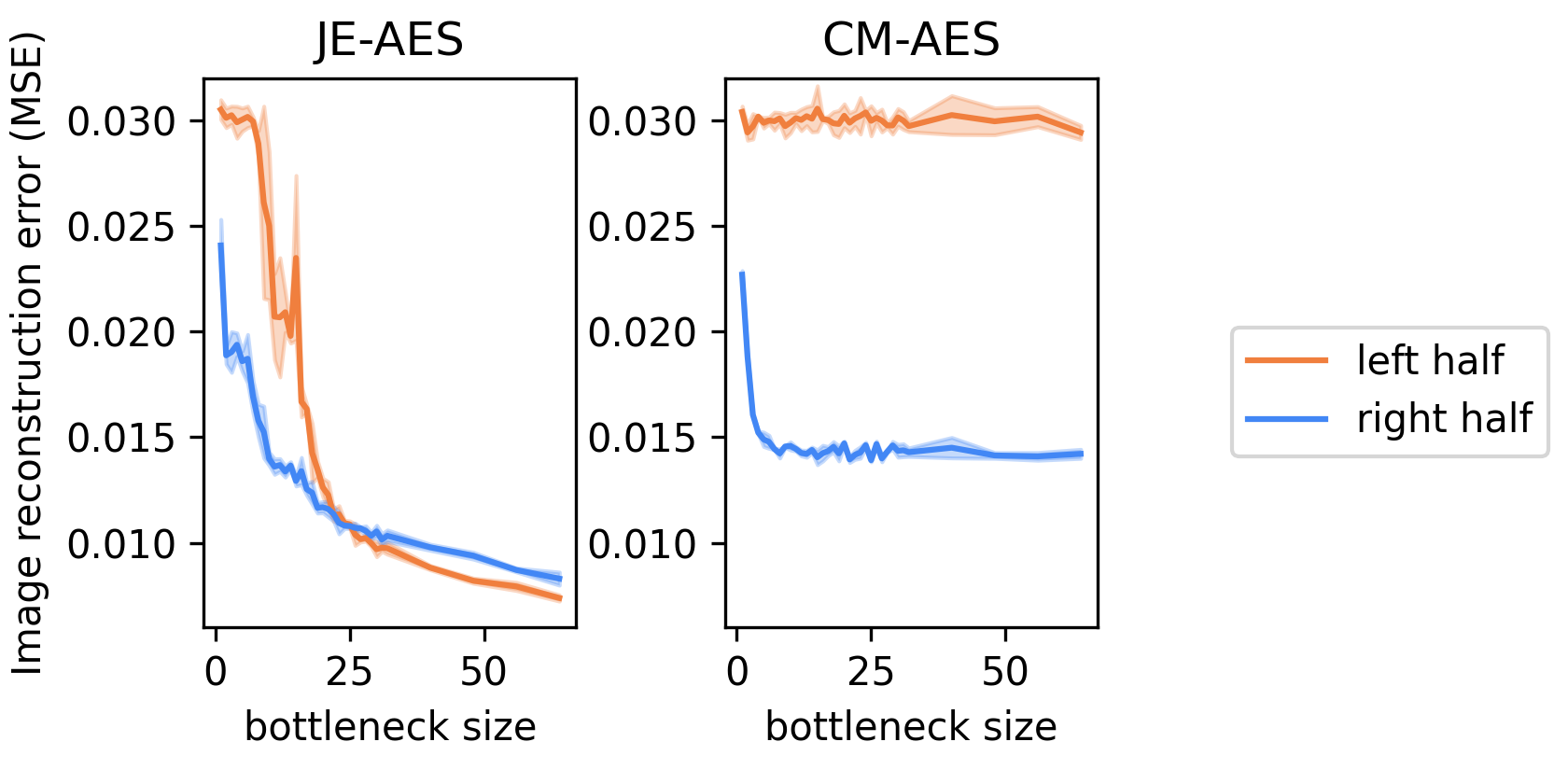}
    \caption{Reconstruction error of the visual modality {\charlescolor for the JE and CM approaches} as a function of the size of the bottleneck of the encoding. The error is split in two parts corresponding to the left and right halves of the frames. The results show that the pixels which share information with the proprioceptive modality are better reconstructed.}
    \label{fig:real_vision_left_right_option_2}
\end{figure}

\begin{figure}[!htb]
    \centering
    \includegraphics[width=\linewidth]{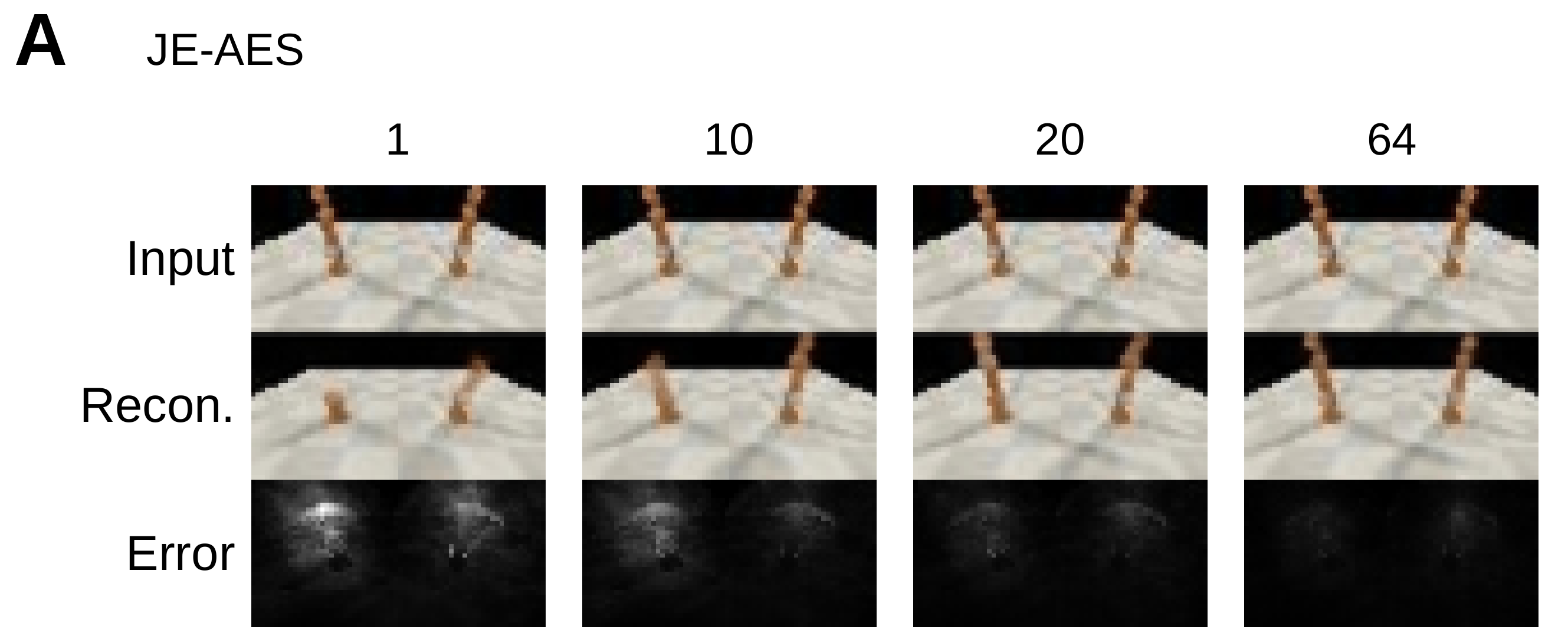}

    \vspace{0.6cm}

    \includegraphics[width=\linewidth]{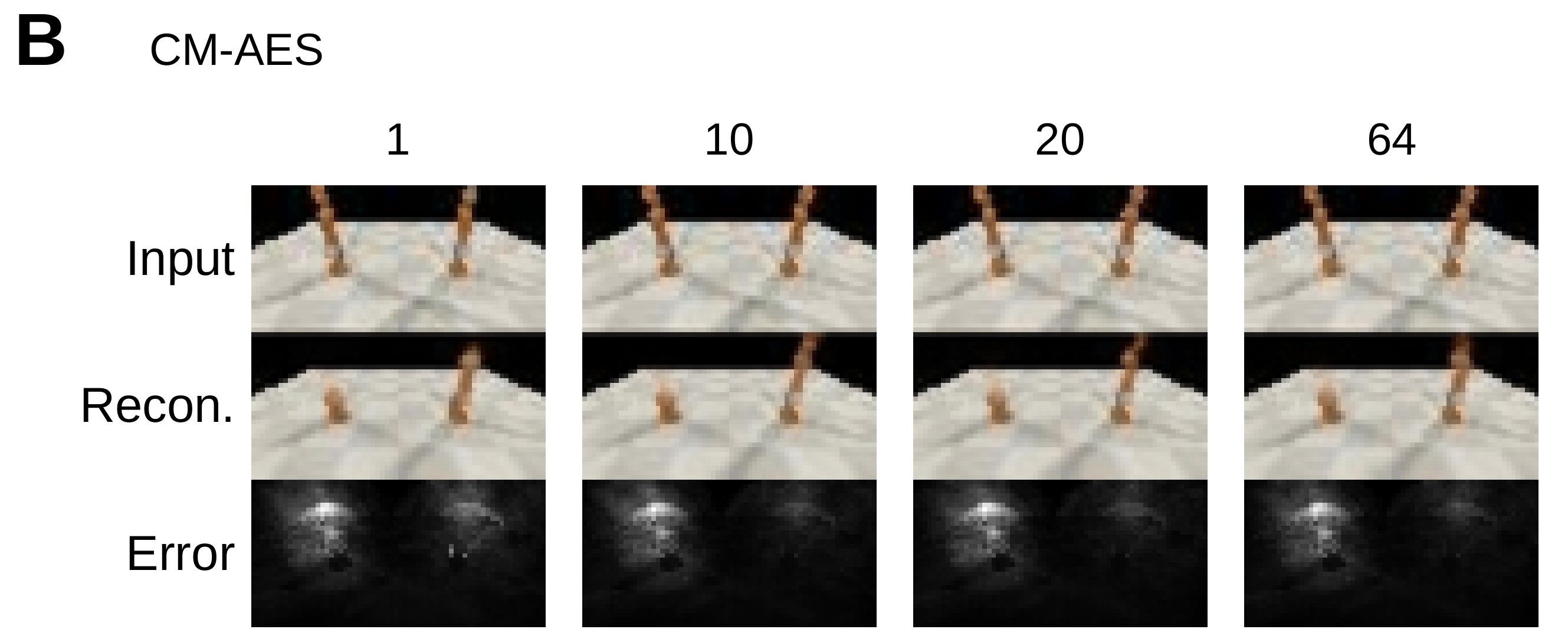}
    \caption{We show for each bottleneck sizes in $\{1, 10, 20, 64\}$ an image and it's reconstruction through the autoencoder, as well as the mean reconstruction error map (darker indicates lower error). {\charlescolor The JE approach (\textbf{A}) encodes the visual information} about both arms if the information bottleneck allows for it, {\charlescolor while the CM approach (\textbf{B}) reconstructs} only one of the two arm for any bottleneck size.}
    \label{fig:real_vision_map_option_2}
\end{figure}

{\revisioncolor 
{\charlescolor We here present the results for the AES option. }
{\charlescolor Overall, the results for the default and the AES options are similar, thus showing that the underlying principle is independent of the implementation details. Compared to the default option, the information about the left arm seems to be marginally better filtered out in the AES option for the CM approach. Figure~\ref{fig:real_readout_option_2} shows the reconstruction error of the various components of the proprioceptive data of both arms as a function of the bottleneck size. Similarly to Fig.~\ref{fig:real_readout_option_1}, the information contained in both the visual and the proprioceptive modalities is better reconstructed than the information present in only one of the modalities. Figure~\ref{fig:real_vision_left_right_option_2} shows the reconstruction error of the data from the vision sensor as a function of the bottleneck size. The two curves correspond to the left and right half of the frames. Similarly to the results presented in Fig.~\ref{fig:real_vision_left_right_option_1}, the part of the image corresponding to the arm which is jointly encoded is better reconstructed than the other. {\charlescolor In the CM approach,} the other arm is not reconstructed at all. Finally, Fig.~\ref{fig:real_vision_map_option_2} shows concrete reconstructions obtained from the encoding process, and the mean reconstruction error map for bottleneck sizes in $\{1, 10, 20, 64\}$.
}
For Figs.~\ref{fig:real_vision_left_right_option_2} and \ref{fig:real_vision_map_option_2}, the frame reconstruction is $B\left(D_\postencoding\left(z, \theta_{D_\postencoding}\right), \theta_B\right)$ with $B$ the decoder part of the autoencoder learning the latent representation $y_\vision$.
}
\ifCLASSOPTIONcompsoc
  \section*{Acknowledgments}
\else
  \section*{Acknowledgment}
\fi

This work was supported by the European Union’s Horizon 2020 Research and Innovation Program under Grant Agreement No 713010 (GOAL-Robots – Goal-based Open-ended Autonomous Learning Robots). JT acknowledges support from the Johanna Quandt foundation.

\ifCLASSOPTIONcaptionsoff
  \newpage
\fi



%

%

\begin{IEEEbiography}[{\includegraphics[width=1in,height=1.25in,clip,keepaspectratio]{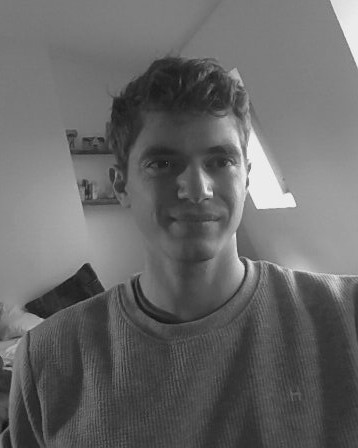}}]{Charles Wilmot} attended preparatory classes for France's \textit{Grandes Ecoles} at the University of Valenciennes and recieved the B.E. in maths in 2013, and received an engineering diploma and the M.E. in information technologies at the \textit{École nationale supérieure d'électronique, informatique, télécommunications, mathématiques et mécanique de Bordeaux} in 2016. He then integrated the research team of Jochen Triesch in Frankfurt, where he studied developmental robotics, intrinsically motivated reinforcement learning and hierarchical reinforcement learning in the intent of obtaining the Ph.D. in 2021.
\end{IEEEbiography}

\begin{IEEEbiography}[{\includegraphics[width=1in,height=1.25in,clip,keepaspectratio]{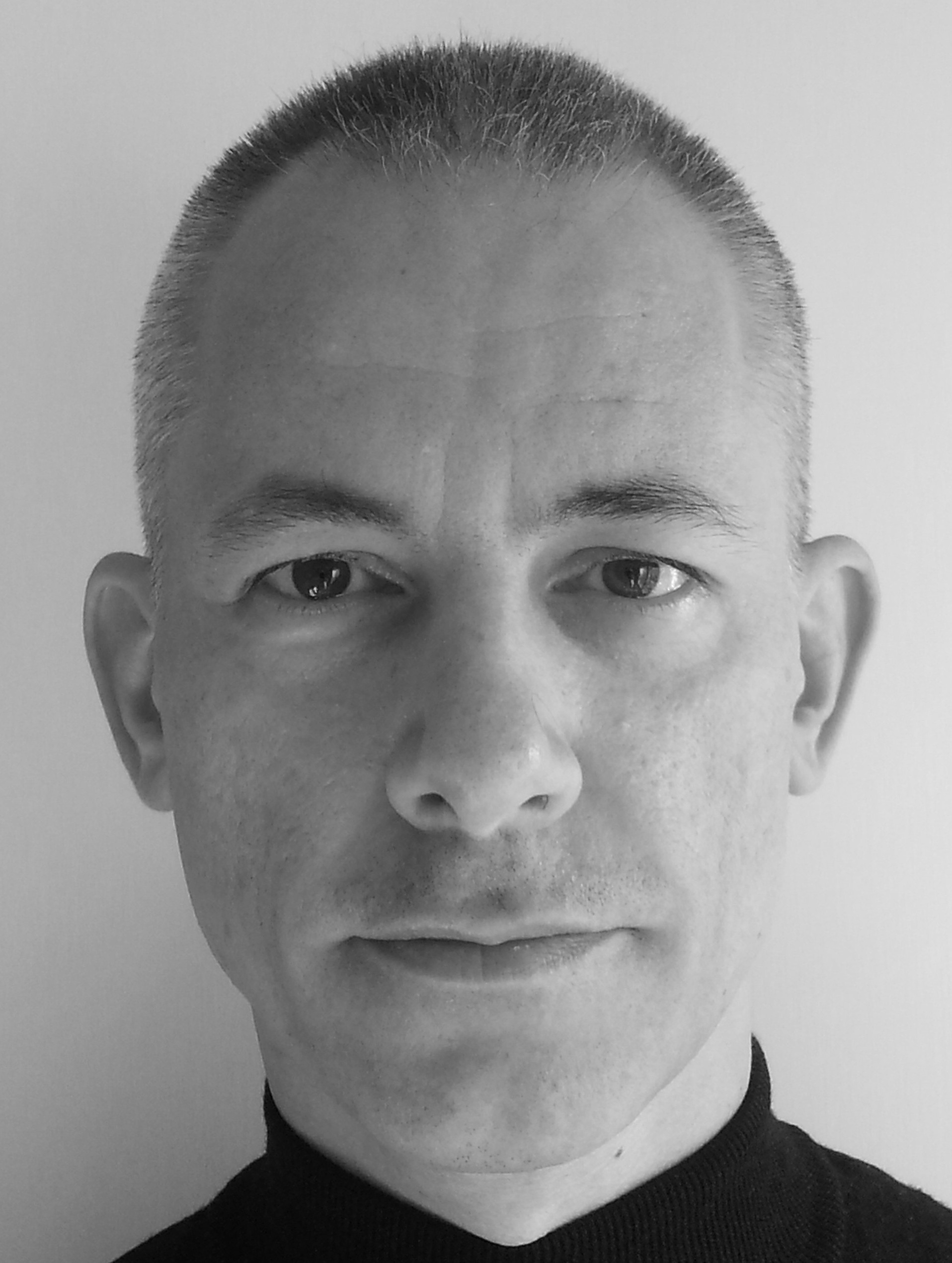}}]{Gianluca Baldassarre} received the B.A. and M.A. degrees in economics from the Sapienza University of Rome, Italy, in 1998, the Diploma of the Specialization Course ``Cognitive Psychology and Neural Networks'' from the same University, in 1999, and the Ph.D. degree in Computer Science from the University of Essex, Colchester, U.K., in 2003. He was later a postdoc with the Italian Institute of Cognitive Sciences and Technologies, National Research Council (ISTC-CNR), Rome. Since 2006 he has been a researcher, now a Research Director, with ISTC-CNR where he founded and is Coordinator of the Research Group ``Laboratory of Embodied Natural and Artificial Intelligence''. He was the Principal Investigator for the EU project ``ICEA--Integrating Cognition Emotion and Autonomy'' from 2006 to 2009, the Coordinator of the EU Integrated Project ``IM-CLeVeR--Intrinsically Motivated Cumulative Learning Versatile Robots'' from 2009 to 2013, and, since 2016, he has been the Coordinator of the EU FET-OPEN Project ``GOALRobots--Goal-Based Open-ended Autonomous Learning Robots'' ending in 2021. His research interests span open-ended learning of sensorimotor skills, driven by extrinsic and intrinsic motivations, in animals, humans, and robots.
\end{IEEEbiography}

\begin{IEEEbiography}[{\includegraphics[width=1in,height=1.25in,clip,keepaspectratio]{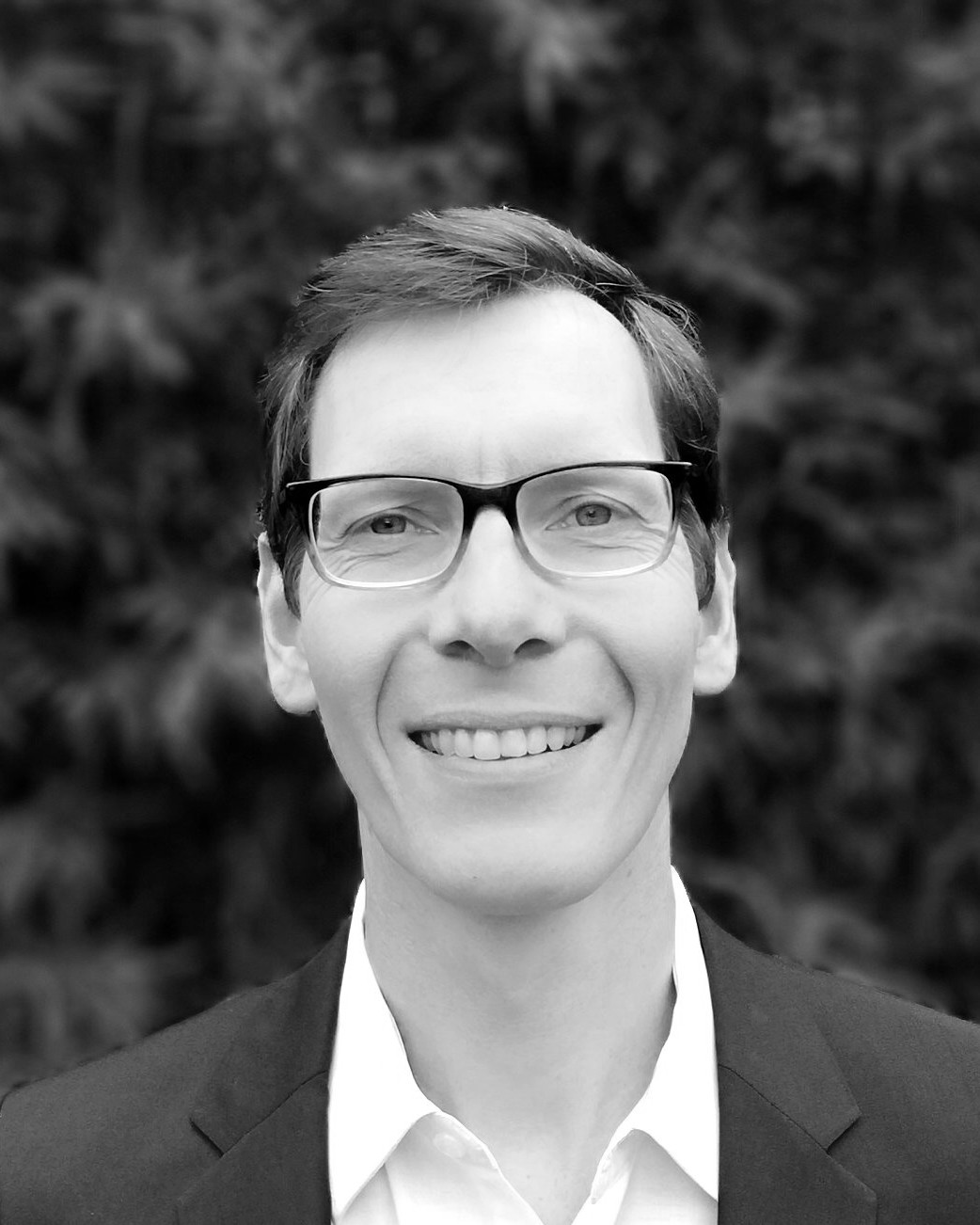}}]{Jochen Triesch} received his Diploma and Ph.D. degrees in Physics from the University of Bochum, Germany, in 1994 and 1999, respectively. After two years as a post-doctoral fellow at the Computer Science Department of the University of Rochester, NY, USA, he joined the faculty of the Cognitive Science Department at UC San Diego, USA as an Assistant Professor in 2001. In 2005 he became a Fellow of the Frankfurt Institute for Advanced Studies (FIAS), in Frankfurt am Main, Germany. In 2006 he received a Marie Curie Excellence Center Award of the European Union. Since 2007 he is the Johanna Quandt Research Professor for Theoretical Life Sciences at FIAS. He also holds professorships at the Department of Physics and the Department of Computer Science and Mathematics at the Goethe University in Frankfurt am Main, Germany. In 2019 he obtained a visiting professorship at the Université Clermont Auvergne, France. His research interests span Computational Neuroscience, Machine Learning, and Developmental Robotics.
\end{IEEEbiography}






\end{document}